\documentclass[a4paper,fleqn]{cas-dc}

\usepackage[numbers]{natbib}
\usepackage[linesnumbered,ruled,vlined]{algorithm2e}
\usepackage{float}
\usepackage[caption = false]{subfig}
\usepackage{orcidlink}

\newcommand{\orcid}[1]{\href{https://orcid.org/#1}{\textcolor[HTML]{A6CE39}{\aiOrcid}}}

\SetCommentSty{mycommfont}

\SetKwInput{KwInput}{Input}                
\SetKwInput{KwOutput}{Output}             

\def\tsc#1{\csdef{#1}{\textsc{\lowercase{#1}}\xspace}}
\tsc{WGM}
\tsc{QE}
\tsc{EP}
\tsc{PMS}
\tsc{BEC}
\tsc{DE}

\begin{document}\sloppy
\let\WriteBookmarks\relax
\def\floatpagepagefraction{1}
\def\textpagefraction{.001}
\shorttitle{Probabilistic Multimodal Depth Estimation}

\title [mode = title]{Probabilistic Multimodal Depth Estimation Based on Camera-LiDAR Sensor Fusion}          

\tnotetext[1]{This document is a result of the research project 17INTER-297, funded by Universidad Autonoma de Occidente.}

\tnotetext[2]{The author undertook this work while he was part of Universidad Autónoma de Occidente}

\author[3]{Johan S. Obando-Ceron\orcidlink{0000-0002-6608-5401}}[style=chinese]
\cormark[2]
\ead{jobando0730@gmail.com}

\author[1]{Victor Romero-Cano\orcidlink{0000-0003-2910-5116}}[style=chinese]
\cormark[1]
\ead{varomero@uao.edu.co}

\author[2]{Sildomar Monteiro\orcidlink{0000-0001-7694-9536}}[style=chinese]
\ead{sildomar@ieee.org}

\address[1]{Robotics and Autonomous Systems Laboratory, Faculty of Engineering, Universidad Autónoma de Occidente, Cali, Valle del Cauca, Colombia}
\address[2]{Autonomy Research Division, Aurora Flight Sciences, Cambridge, MA, USA}
\address[3]{University of Montreal, Montréal, Quebec, Canada}

\begin{abstract}
   Multi-modal depth estimation is one of the key challenges for endowing autonomous machines with robust robotic perception capabilities.
There have been outstanding advances in the development of uni-modal depth estimation techniques based on either monocular cameras, because of their rich resolution, or LiDAR sensors, due to the precise geometric data they provide.
However, each of these suffers from some inherent drawbacks, such as high sensitivity to changes in illumination conditions in the case of cameras and limited resolution for the LiDARs.
Sensor fusion can be used to combine the merits and compensate for the downsides of these two kinds of sensors.
Nevertheless, current fusion methods work at a high level.
They process the sensor data streams independently and combine the high-level estimates obtained for each sensor.
In this paper, we tackle the problem at a low level, fusing the raw sensor streams, thus obtaining depth estimates which are both dense and precise, and can be used as a unified multi-modal data source for higher level estimation problems.
   
This work proposes a Conditional Random Field model with multiple geometry and appearance potentials. It seamlessly represents the problem of estimating dense depth maps from camera and LiDAR data.
The model can be optimized efficiently using the Conjugate Gradient Squared algorithm.
The proposed method was evaluated and compared with the state-of-the-art using the commonly used KITTI benchmark dataset.

\end{abstract}


\maketitle

\section{Introduction}

Autonomous robots are composed of different modules that allow them to perceive, learn, decide and act within their environment.
The perception module processes cues that inform the robot about the appearance and geometry of the environment.
Especially when working in outdoor or underground scenarios, these cues must be robust to unseen phenomenon.
A fully autonomous robot must execute all operations, monitor itself, and be able to handle all unprecedented events and conditions, such as unexpected objects and debris on the road, unseen environments, adverse weather, etc.
Therefore, reliable and robust perception of the surrounding environment is one of the key tasks of autonomous robotics.

Among the main inputs for a perception system are the distances of the robot from multiple points in its environment.
This input can be obtained directly from a sensor or estimated by a depth estimation module.
Depth estimation can be performed by processing monocular camera images, stereo vision, radar or LiDAR (Light Detector and Ranging) sensors, among others.
Although monocular cameras can only be used to generate depth information up-to-a-scale, they are still an important component of a depth estimation system due to their low price and the rich appearance data they provide.

Although a monocular camera is small, low-cost and energy efficient, it is very sensitive to changes in the illumination.
Additionally, the accuracy and reliability of depth estimation methods based on monocular images is still far from being practical. For instance, the state-of-the-art RGB-based depth prediction methods \cite{A2, A3, A4} produce an average error (measured by the root mean squared error) of over 50 cm in indoor scenarios (e.g., on the NYU-Depth-v2 dataset \cite{A5}).
Such methods perform even worse outdoors, with at least 4 meters of average error on the Make3D and KITTI datasets \cite{A6, A7}.

3D LiDAR scanners, on the other hand, can provide accurate geometric information about the environment even when illumination changes occur.
Even though, due to their active nature, LiDARs are robust to dark or overexposed scenarios, the generated 3D point cloud is sparse and non-uniformly distributed, which decreases its utility for recognition tasks.
In general, each type of sensor has its own weaknesses.

To address the potential fundamental limitations of image-based depth estimation, this paper considers the use of sparse depth measurements, along with RGB data, to reconstruct depth in full resolution.
Data fusion techniques have been extensively employed for robust perception systems, where fusing and aggregating data from different sensors is required. Although some approaches to robust perception resort to statistical methods for dealing with data outliers \cite{A43}, the work presented in this paper belongs to the group that tackles the robust-perception problem by leveraging the complementary nature of passive and active sensor modalities.

Multi-sensor approaches to robotic perception can be categorised according to the level at which the data from the different sensing modalities is fused in order to obtain the estimate of interest.
According to \cite{A44}, data fusion can be made at the level of symbolic estimates (high level fusion), at the level of features (medium level fusion), or at the level of raw data (low level fusion).

In this paper, a low level fusion method is developed.
It explores the complementary relations between the passive and active sensors at the pixel level \cite{A48, A72, ceron2019probabilistic}.
The approaches in \cite{A45, A46, A47} follow this intuition but require the fused modalities to have similar coverage densities.
Our proposed framework provides a procedure for fusing LiDAR and image data independently of the LiDAR data’s density.

The main contribution of this paper is a depth regression model that takes both a sparse set of depth samples and RGB images as the inputs, and predicts a full-resolution depth map.
This is achieved by modelling the problem of fusing low resolution depth images with high resolution camera images as a Conditional Random Field (CRF).

The intuition behind our CRF formulation is that depth discontinuities in a scene often co-occur with changes in colour or brightness within the associated camera image.
Since the camera image is commonly available at much higher resolution, this insight can be used to enhance the resolution and accuracy of the depth image.
A depth map will be produced by our approach using three features as illustrated in \textcolor{blue}{Fig.~1}.
The first one is an RGB colour image from the camera sensor, top image \textcolor{blue}{Fig.~1 (a)}.
The second one is 2D sparse depth map captured by a LiDAR sensor, middle image \textcolor{blue}{Fig.~1 (b)}.
The third feature is a surface normal map generated from
the sparse depth samples, bottom image \textcolor{blue}{Fig.~1 (c)}.

The rest of this paper is organized as follows.
Section \ref{RW} reviews related work on depth estimation.
Section \ref{ILCF} explains how the LiDAR points and the camera images were registered.
In Section \ref{CRF}, we first introduce our CRF-Fusion framework, then we provide a detailed explanation of the proposed model: the energy potentials that compose our CRF model and its inference machine.
The experimental validation, performed on the KITTI dataset, is reported in Section \ref{RD}.
Finally, conclusions and directions for future work are listed in Section \ref{CFW}.

\begin{figure}
	\centering
	\includegraphics[scale=.187]{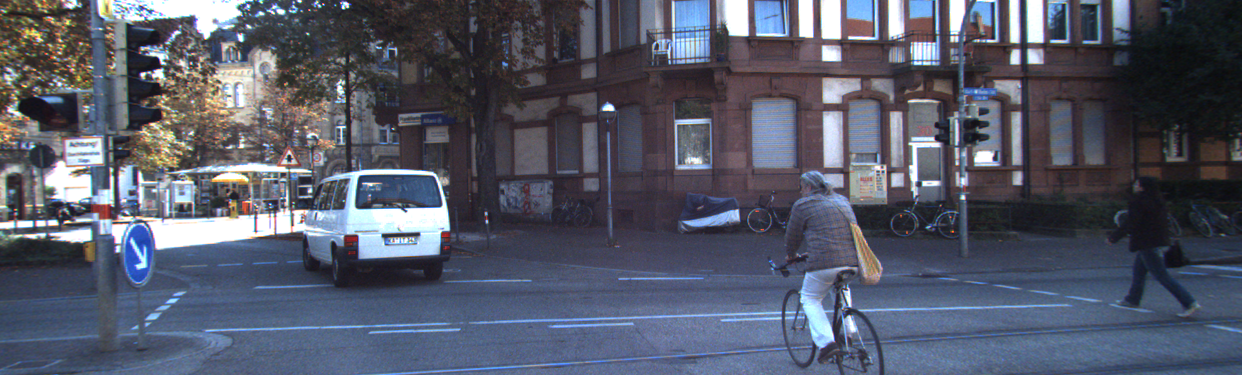}
	\includegraphics[trim={0.8cm 0 0 0},clip,scale=.236]{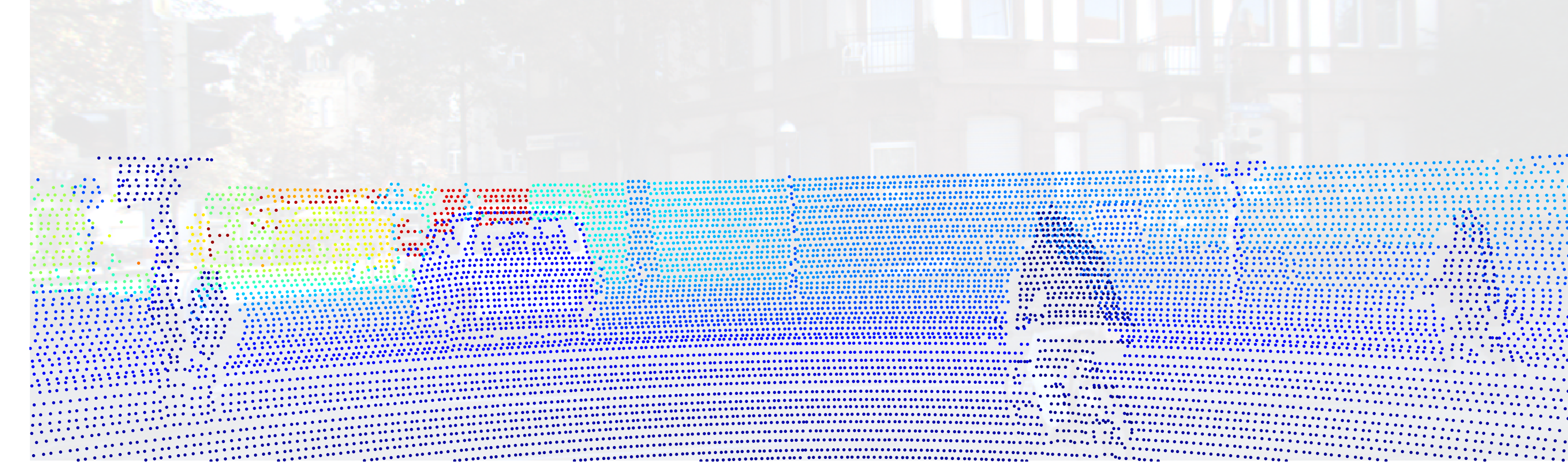}
    \includegraphics[trim={0.8cm 0 0 0},clip,scale=.233]{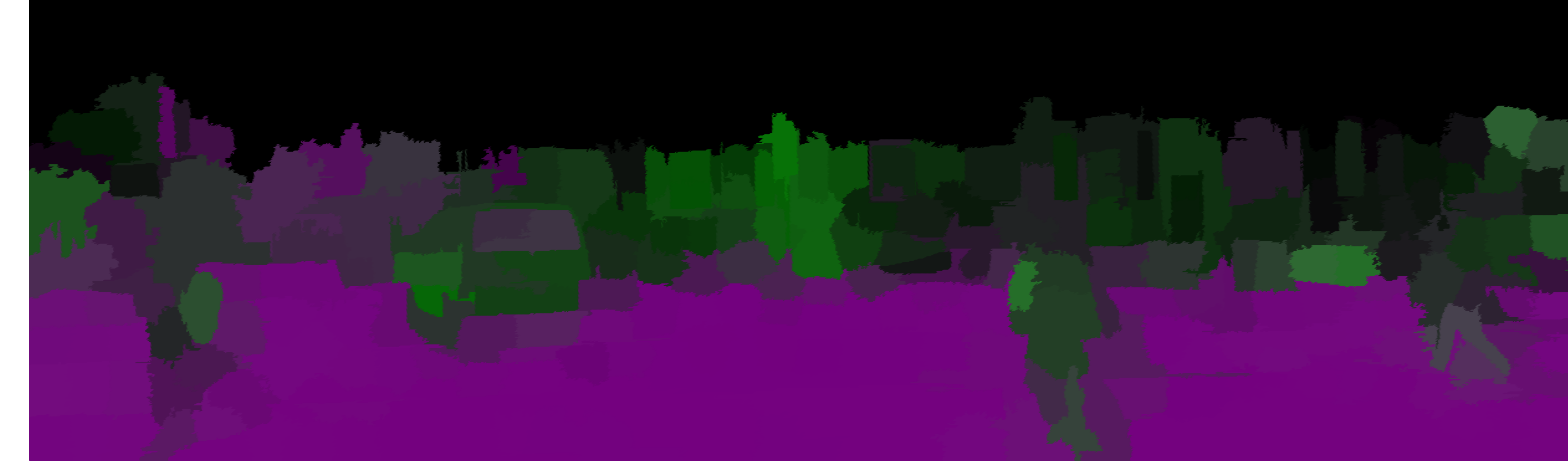}

	\caption{Input features of our framework. We developed a CRF regression model to predict a dense depth image from a single RGB image, and a set of sparse depth samples:  (a), (b) and (c) are the input RGB image, a set of sparse depth samples projected on the image plane, and the projected surface normals, respectively}
	\label{FIG:1}
\end{figure}

\section{Related work}
\label{RW}

Depth estimation from monocular images is a long-standing problem in computer vision.
Early works on depth estimation using RGB images usually relied on hand-crafted features and inference on probabilistic graphical models.
 Classical methods include shape-from-shading \cite{A8} and shape-from-defocus \cite{A9}.
Other early methods were based on hand-tuned models or assumptions about the orientations of the surfaces \cite{A10}.
Newer methods treat depth estimation as a machine learning problem, most recently using deep artificial neural networks \cite{A11, A12}.
For instance, Saxena et al. \cite{A13} estimated the absolute scales of different image patches and inferred a depth image using a Markov Random Field model.
Eigen et al. used a multiscale convolutional network to regress from colour images to depths \cite{A11, A3}.
Laina et al. used a fully convolutional network architecture based on ResNet \cite{A4}.
Liu et al. proposed a deep convolutional neural field model that combines deep networks with Markov random fields \cite{A17}. Roy et al. combined shallow convolutional networks with regression forests to reduce the need for large training sets \cite{A18}. In \cite{A19} the proposed attention model is seamlessly integrated with a CRF, allowing end-to-end training of the entire architecture.
This approach benefits from a structured attention model which automatically regulates the amount of information transferred between corresponding features at different scales.

The approach of Li et al. \cite{A20} combines deep learning features on image patches with hierarchical CRFs defined on a superpixel segmentation of the image.
They use pretrained AlexNet \cite{A21} features of image patches to predict depth at the centre of the superpixels.
A hierarchical CRF refines the depth across individual pixels.
Liu et al. \cite{A2} also propose a deep structured learning approach that avoids hand-crafted features.
They presented a deep structured learning scheme which learns the unary and pairwise potentials of a continuous CRF in a unified deep CNN framework.
Liu et al. \cite{A23} proposed a discrete-continuous CRF model to take into consideration the relations between adjacent superpixels, e.g., occlusions.

Recent work has also shown the benefit of adopting multi-task learning strategies, e.g. for jointly predicting depth and performing semantic segmentation, ego-motion estimation or surface normal computation \cite{A25, A26}. Some recent papers have proposed unsupervised or weakly supervised methods for reconstructing depth maps \cite{A27, A28}.

With  the rapid development of deep neural networks, monocular depth estimation based on deep learning and computer vision tecniques has been widely studied recently and  achieved  promising  performance  in terms of accuracy \cite{A81}.
 However, not considering information from other sensors makes the estimate not so robust.
In the mentioned literature, there are used different kinds of network frameworks, loss functions, and training strategies with just one sensory modality.
The architecture proposed in this paper uses two sensory modalities.

Fusing data coming from multiple sensors has the potential to improve the robustness of the depth estimates.
Ma et al. \cite{A29} uses RGB images together with sparse depth information to train a bottleneck network architecture.
Compared to imagery-only methods, their approach generates better depth estimation results.
Others have investigated depth estimation from colour images augmented with sparse sets of depth measurements using probabilistic graphical models.
The techniques described in \cite{A30, A31, A32}, and \cite{A33} are able to fuse the information from both sources to significantly improve the resolution of low quality and sparse range images.

Wang et al. proposed a multi-scale feature fusion method for depth completion \cite{A34} using sparse LIDAR data.
Ma et al. proposed two methods: a supervised method for depth completion using a ResNet based architecture and a self-supervised method which uses the sparse LiDAR input along with pose estimates to add additional training information based on depth and photometric losses \cite{A36}.

Although recent methods have achieved impressive progress in terms of evaluation metrics such as the pixel-wise relative error, most of them neglect the geometric constraints in 3D space.
This component is considered in our CRF model, which makes this approach different from previous fusion methods.

Providing  strong cues from surface information is relevant for improving the accuracy  of the depth prediction \cite{A63}.
Recently, Zhang et al. \cite{A61} proposed predicting surface normals and  occlusion  boundaries using a  deep  network  and  further used them to help depth completion in indoor scenes.
The works in \cite{A62,Xu2019} propose end-to-end deep learning systems to produce dense depth maps from sparse LiDAR data and a colour image taken from outdoor on-road scenes, leveraging surface normals as the intermediate representation. Zhang et al. \cite{A61} predicted surface normals by leveraging RGB data,  leading to a better prior for depth completion.
They ultimately combined these predictions with  sparse depth input to generate a complete depth map.
Our method takes advantage of the surface normals to improve the performance of the proposed model. \cite{Xu2019}, in particular, uses a diffusion layer to refine the completions. \cite{Liu2023} shows that diffusion models can be interpreted as energy-based models (EBMs). Now, CRFs can also be interpreted as energy-based models parameterized as factor graphs. Therefore, there is a relation between the diffusion and the CRF approach as both can be interpreted as EBMs, but with different formulations and assumptions. For the problem of depth estimation, the CRF approach more naturally captures the prior knowledge of the geometric constraints inherent in the physics underlying the LiDAR data measurements, whereas the diffusion approach relies on a carefully crafted similarity metric, which might make the models unstable and hard to train. In other words, we expect a CRF approach to require less training to achieve equivalent performance as the diffusion-based method.

The problem of fusing LiDAR and image data can be approached as a camera pose estimation problem, where the relation between the 3D LIDAR coordinates and the 2D image coordinates is characterised by camera parameters, such as the position, orientation, and focal length.
In \cite{A37}, there was proposed an information-theoretic similarity measure to automatically register 2D-Optical imagery with 3D LiDAR scans by searching for a suitable camera transformation matrix.
The fusion of 3D-LiDAR data with stereoscopic images is addressed in \cite{A38, A39, A40, A80}.
The advantage of stereoscopic depth estimation is its ability to produce dense depth maps of the surroundings by using stereo matching techniques.
In \cite{A80}, for example, there was proposed a two-stage cascade deep architecture that first fuses stereo and LiDAR disparity maps and then refines the estimated disparity maps by introducing colour features.
In contrast, our method does not rely on a stereo matching algorithm, something which tends to be computationally costly.

\section{Image and LIDAR point cloud registration}
\label{ILCF}
This section gives a brief introduction to the process of aligning an image and a LIDAR point cloud, which allows the projection of LiDAR points onto the image plane.
As presented in \cite{A41}, in a robotic platform equipped with both a LiDAR and a camera, these two sensors are synchronized so simultaneous scans from both sensor are collected.
The camera and LiDAR are cross-calibrated so that the point cloud can be projected onto the image plane \cite{A41}.
Once projected, the LiDAR points are associated with either pixels or groups of pixel, also called superpixels.
This section also briefly describes the Simple Linear Iterative Clustering (SLIC) algorithm, by which these superpixels are obtained. \\

\subsection{Point cloud projection}
In order to fuse the image and LiDAR data, it is imperative to find mathematical models that represent the spatial correspondence of pixels and 3D points.
These models allow us to project the LiDAR point cloud onto the image plane.

The projection $\mathbf{y}$ of a 3D point $\mathbf{x}=(x, y, z, 1)^{T}$ in rectified
and rotated camera coordinates to a point $\mathbf{y}=(u, v, 1)^{T}$ in the $i^{\prime}$th camera image is given by


$$
\mathbf{y}=\mathbf{P}_{\text {rect}}^{(i)} \mathbf{x}
$$

\noindent with 

$$
\mathbf{P}_{r e c t}^{(i)}=\left(\begin{array}{cccc}{f_{u}^{(i)}} & {0} & {c_{u}^{(i)}} & {-f_{u}^{(i)} b_{x}^{(i)}} \\ {0} & {f_{v}^{(i)}} & {c_{v}^{(i)}} & {0} \\ {0} & {0} & {1} & {0}\end{array}\right)
$$

\noindent being the $i$th projection matrix.
Here, $b_{x}^{(i)}$ denotes the baseline with respect to a reference camera.
Note that in order
to project the 3D point $\mathbf{x}$, in the camera reference coordinates, to a point $\mathbf{y}$ on the $i$th image plane, the rectifying rotation matrix of the reference camera $\mathbf{R}_{r e c t}^{(0)}$ must be considered as well.

$$
\mathbf{y}=\mathbf{P}_{r e c t}^{(i)} \mathbf{R}_{r e c t}^{(0)} \mathbf{x}
$$

Here, $\mathbf{R}_{r e c t}^{(0)}$ has been expanded into a 4×4 matrix by appending a fourth zero-row and column, and setting $\mathbf{R}_{r e c t}^{(0)}(4,4)=1$.
We also need to register the laser scanner with respect to the camera's coordinate system.
The rigid-body transformation from LiDAR coordinates to camera coordinates is given by

$$
\mathbf{T}_{v e l o}^{c a m}=\left(\begin{array}{cc}{\mathbf{R}_{v e l o}^{\operatorname{cam}}} & {\mathbf{t}_{v e l o}^{c a m}} \\ {0} & {1}\end{array}\right)
$$

Finally, a 3D point $\mathbf{x}$ in the LiDAR coordinate system gets projected to a point $y$ in the $i$th camera image:

$$
\mathbf{y}=\mathbf{P}_{r e c t}^{(i)} \mathbf{R}_{r e c t}^{(0)} \mathbf{T}_{v e l o}^{c a m} \mathbf{x}
$$

Subsequently, as a preprocessing step, points with a negative value of $z$ are removed.
Then the remaining points can be projected onto the image
plane using the projection matrix

$$\left[x^{\prime} y^{\prime} z^{\prime}\right]^{T}=\mathbf{y}\left[x_{p}\;y_{p}\; z_{p}\; 1\right]^{T}$$


The projected pixel coordinates of the LIDAR points can be obtained by

$$[x, y]=\left[\frac{x^{\prime}}{z^{\prime}}, \frac{y^{\prime}}{z^{\prime}}\right]$$


\textcolor{blue}{Fig.~2} portrays the projection of a point cloud projection onto the image plane.

\begin{figure}
	\centering
		\includegraphics[trim={0.8cm 0 0 0},clip,scale=.235]{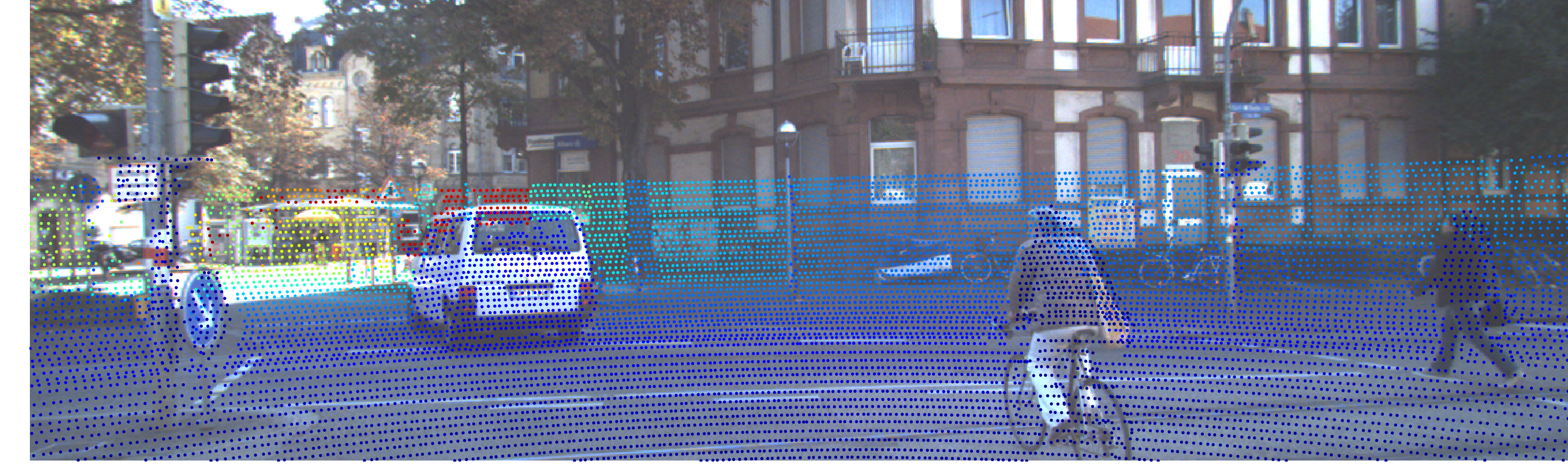}

		\includegraphics[trim={0.8cm 0 0 0},clip,scale=.30]{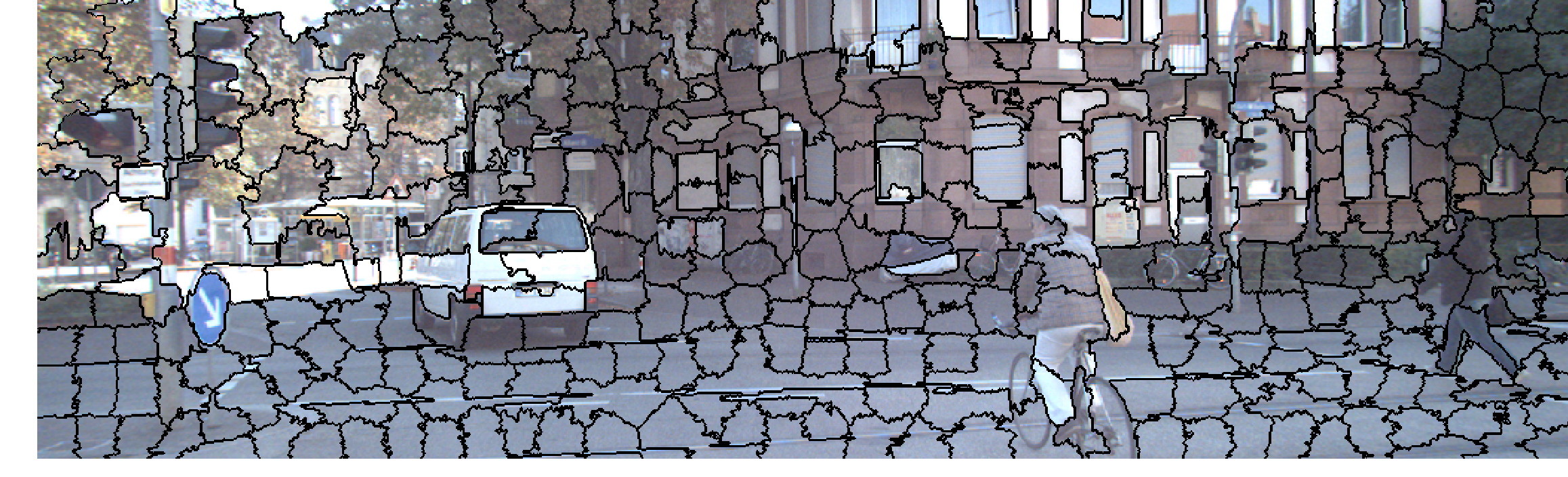}
		
		\includegraphics[trim={0.8cm 0 0 0},clip,scale=.30]{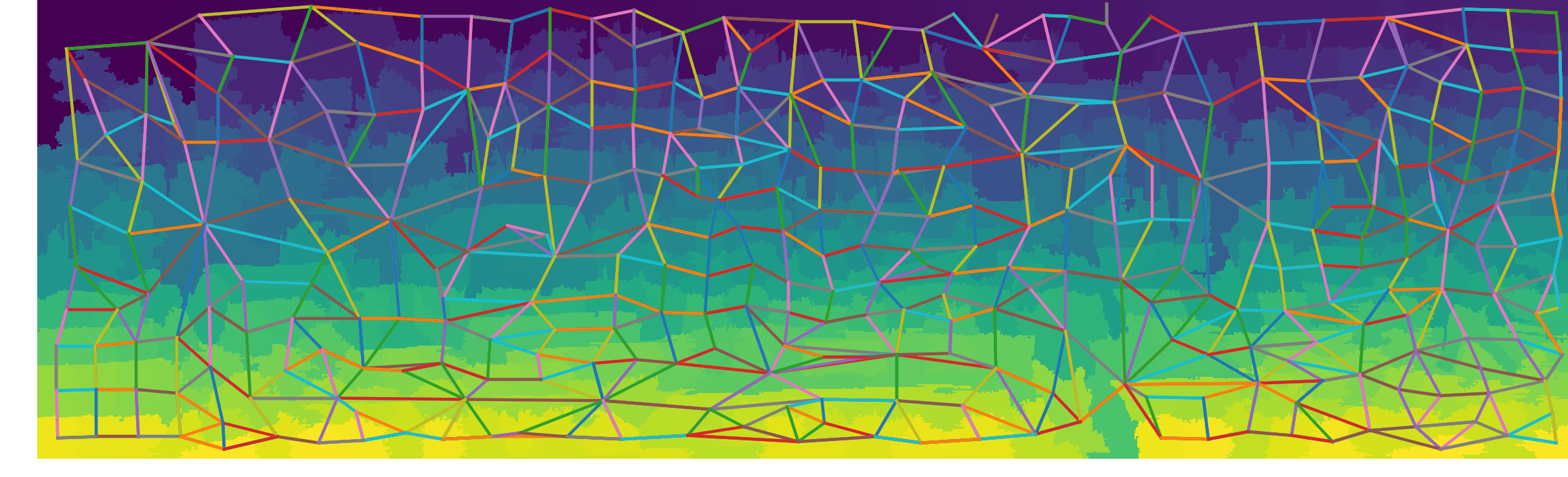}
		
		\includegraphics[trim={0.8cm 0 0 0},clip,scale=.301]{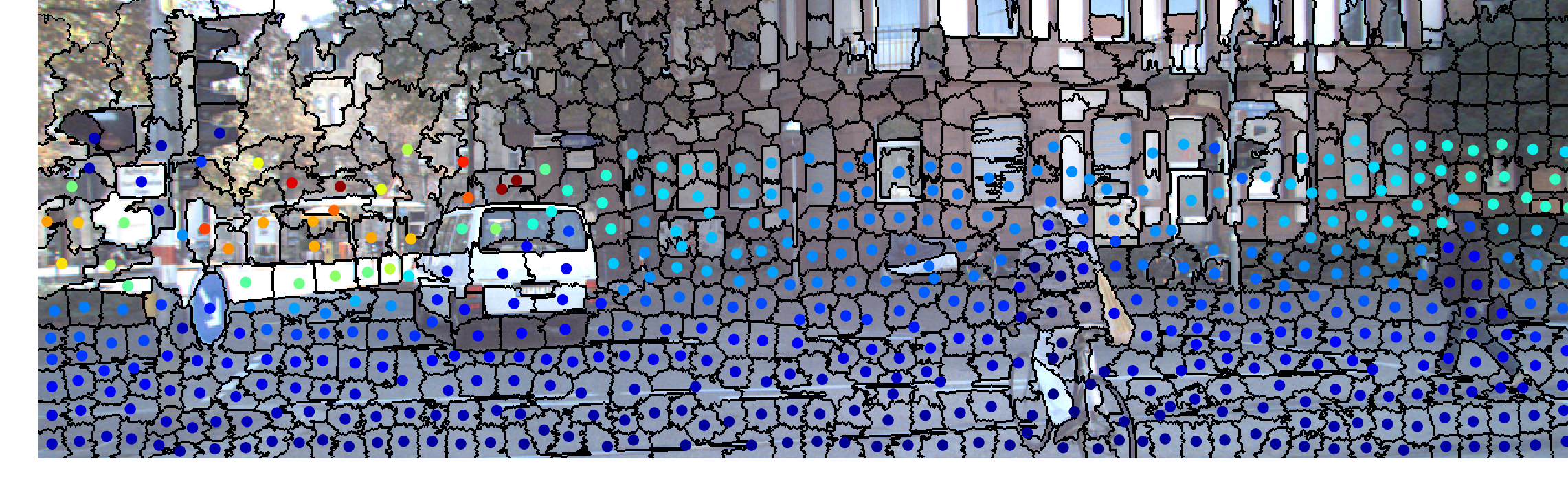}

	\caption{ Given an (a) input image and its corresponding depth, where dark blue indicates closer distances, our work is focused on densifying the input sparse depth.
We combine the (b) superpixel segmentation with the sparse input depth to obtain our initial superpixel depth.
We use a CRF to enhance the resolution and accuracy of the depth image considering information from the 4 neighbours for each super pixel.
	In (d) we can see how we achieve significant improvements with regard to the sparse input  depth.
Each pixel inside the superpixels has the same depth value.
	}
	\label{FIG:2}
\end{figure}

\subsection{Superpixel segmentation using Simple Linear Iterative Clustering (SLIC)}

Considering superpixels rather than individual pixels greatly reduces the complexity of the subsequent image processing  tasks.
In order to harness the full potential of the use of superpixels, their calculation must be fast, easy to use, and  produce  high  quality  segmentations.

Superpixel segmentation algorithms provide  an  effective  way to extract information about local image features.
Our framework uses an implementation of the SLIC algorithm to group sparse depth measurements which have been previously projected onto the image plane.
SLIC is a simple and parallelisable pixel clustering method, based on the $k$-means algorithm, which is used for decomposing an image into a regular grid of visually homogeneous regions or so-called superpixels \cite{A42}.
As a result, SLIC superpixels provide a regular grouping of image pixels according to their distance both spatially and in the colour space.
We use this superpixel segmentation method to assign depth values from a sparse point cloud to all of the pixels within the superpixels.

Applying SLIC to the original image provides a segmentation with as many segments as the number of superpixels set as a hyperparameter.
Each segment is identified by an ID that allows individual pixels to be assigned to superpixel segments.
Additionally, the coordinates of the segment's centroid and those of its neighbours are also output by a super-segmentation step.
Thanks to the superpixel segmentation, each node in our proposed CRF will be associated with a small number of segments, rather than have the extremely large quantity of individual pixels usually present in high definition images.

As provided by the super-segmentation step, the number of neighbours around a segment may vary, depending on the spatial homogeneity of the colour in the image and the expected number of segments.
After segmentation, the grid-like structure of the original images is lost.

The grid-structure of a CRF requires that each node, and therefore each segment, be associated with only four neighbours.
Thus it is necessary to find the four nearest neighbours corresponding to each segment, since the grid-like structure is not well defined.

\begin{figure}
	\centering
		\includegraphics[trim={0 0 0 0},clip,scale=.38]{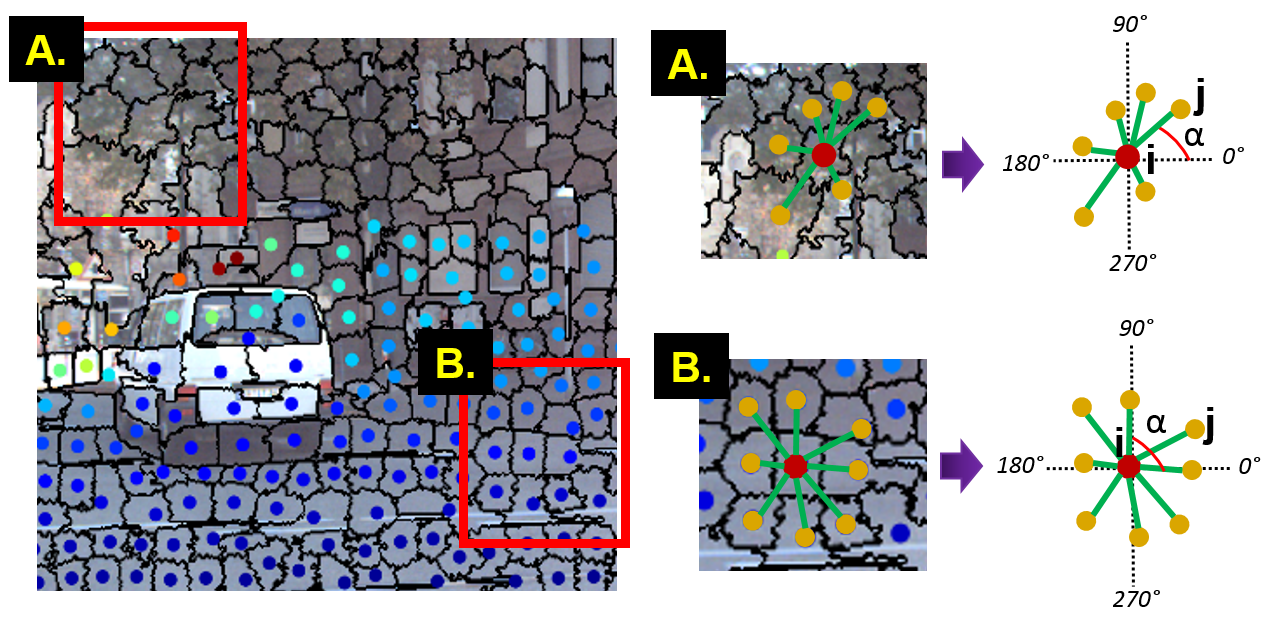}

	\caption{Two examples of superpixels and their neighbours.
The red dots in A and B are nodes assigned to two different super pixels, while the green lines represent their corresponding nearest neighbours.
Note that the super segmentation used allows a superpixel node to have more than four neighbours.}
		
	\label{FIG:3}
\end{figure}

In order to determine which superpixels are the four closest neighbours, the angles between the centroid of each segment and its neighbours are calculated.
We select as neighbours the four segments whose angles have the least difference from 0, 90, 180 and 270 degrees, respectively.
For two examples of nodes, depicted as red points, \textcolor{blue}{Fig.~\ref{FIG:3}} illustrates their closest neighbours, whose centroids are represented by yellow dots.

\textcolor{blue}{Fig.~4} shows the 4 closest neighbours selected.
These neighbours are represented by dark green lines and dark yellow dots.
Note that for the superpixels located at the corners, respectively, on the edges, only the 2, respectively, 3, closest neighbours need to be found.

\begin{figure}
	\centering
		\includegraphics[trim={0 0 0 0},clip,scale=.3]{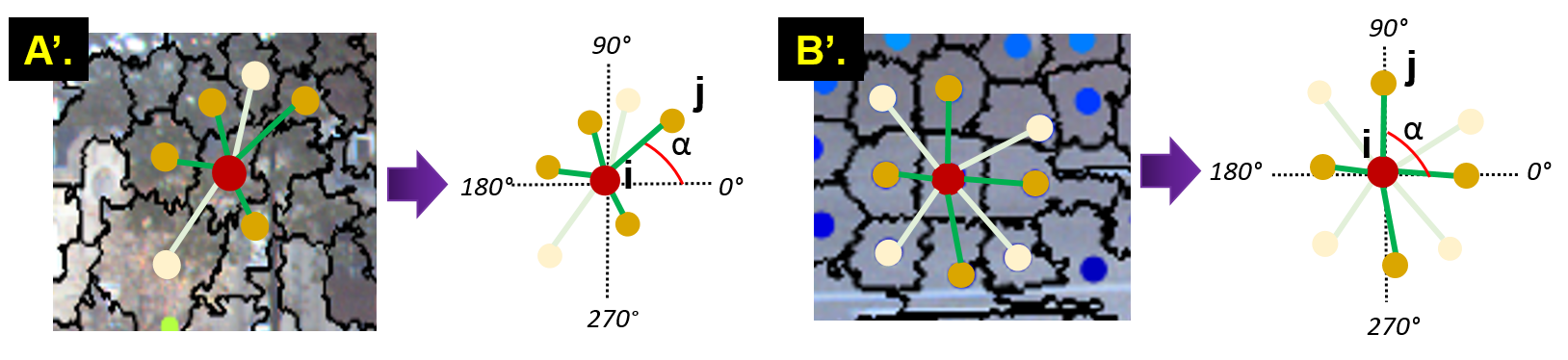}

	\caption{Selected 4 nearest neighbours for super pixel nodes (red dots) in Fig. \ref{FIG:3}.
Dark green lines connect nodes with their selected neighbours.}
		
	\label{FIG:4}
\end{figure}

\section{CRF-based camera-LIDAR fusion for depth estimation}
\label{CRF}

In this paper, depth estimation is formulated as a superpixel-level inference task on a modified Conditional Random Field (CRF).
Our proposed model is a multi-sensor extension of the classical pairwise CRF.
In this section, we first briefly introduce the CRF model.
Then we show how to fuse the information of an image and a sparse LIDAR point cloud with our novel CRF framework.
\subsection{Overview}
The Conditional Random Field (CRF) is a type of undirected probabilistic graphical model which is widely used for solving labeling problems.
Formally, let $\mathbf{X}=\left\{X_{1}, X_{2}, \ldots, X_{N}\right\}$ be a set of discrete random variables to be inferred from an observation or input tensor $\mathbf{Y}$, which in turn is composed of the observation variables $c_{i}$ and $y_{i}$, where $i$ is an index over superpixels.
For each superpixel $i$, the variable $c_{i}$ corresponds to an observed three-dimensional colour value and $y_{i}$ is an observed range measurement.

The goal of our framework is to infer the depth of each pixel in a single image depicting general scenes.
Following the work of \cite{A31, A49} we make the common assumption that an image is composed of small homogeneous regions (superpixels) and consider a graphical model composed of nodes defined on superpixels.
Note that our framework is flexible and can estimate depth values on either pixels or superpixels.

The remaining question is how to parametrize this undirected graph.
Because the interaction between adjacent nodes in the graph is not directed, there is no reason to use a standard Conditional Probability Distribution (CPD), in which one represents the distribution over one node given the others.
Rather, we need a more symmetric parametrization.
Intuitively, we want our model to capture the affinities between the depth estimates of the superpixels in a given neighbourhood.
These affinities can be captured as follows: Let $\tilde{P}(X,Y)$ be an unnormalized Gibbs joint distribution parametrized as a product of factors $\Phi$, where

$$
\Phi=\left\{\phi_{1}\left(D_{1}\right), \ldots, \phi_{k}\left(D_{k}\right)\right\},
$$

\noindent and

$$
\tilde{P}(X,Y)=\prod_{i=1}^{m} \phi_{i}\left(D_{i}\right).
$$

We can then write a conditional probability distribution of the depth estimates $X$ given the observations $Y$ using the previously introduced Gibbs distribution, as follows:

$$
Pr(X | Y)=\frac{P(X, Y)}{Z(Y)}
$$

\noindent where,

$$
Z(Y)=\sum_{X} \tilde{P}(X, Y).
$$

Here, $Z(Y)$, also known as `the partition function', works as a normalizing factor which marginalizes $X$ from $\tilde{P}(X,Y)$, allowing the calculation of the probability distribution $P(X | Y)$:

$$
P(X | Y)=\frac{1}{\sum_{X} \tilde{P}(X, Y)} \tilde{P}(X,Y).
$$

Therefore, similar to conventional CRFs, we model the conditional probability distribution of the data with the following density function:

$$
\operatorname{P}(\mathbf{X} | \mathbf{y})=\frac{1}{\mathrm{Z}(\mathbf{Y})} \exp (-E(\mathbf{X}, \mathbf{Y}))
$$

\noindent where $E$ is the energy function and $Z$ is the partition function defined by

$$
\mathrm{Z}(\mathrm{Y})=\int_{\mathrm{Y}} \exp \{-E(\mathrm{X}, \mathrm{Y})\} \mathrm{d} \mathrm{Y}.
$$

Since $Z$ is continuous, this integral equation can be analytically solved.
This is different from the discrete case, in which approximation methods need to be applied.
To predict the depths of a new image, we solve the following maximum a posteriori (MAP) inference problem:

$$
\mathbf{x}^{\star}=\underset{\mathbf{x}}{\operatorname{argmax}} \operatorname{P}(\mathbf{X} | \mathbf{Y}).
$$

\begin{figure*}
\centering
    \includegraphics[width=0.8\linewidth]{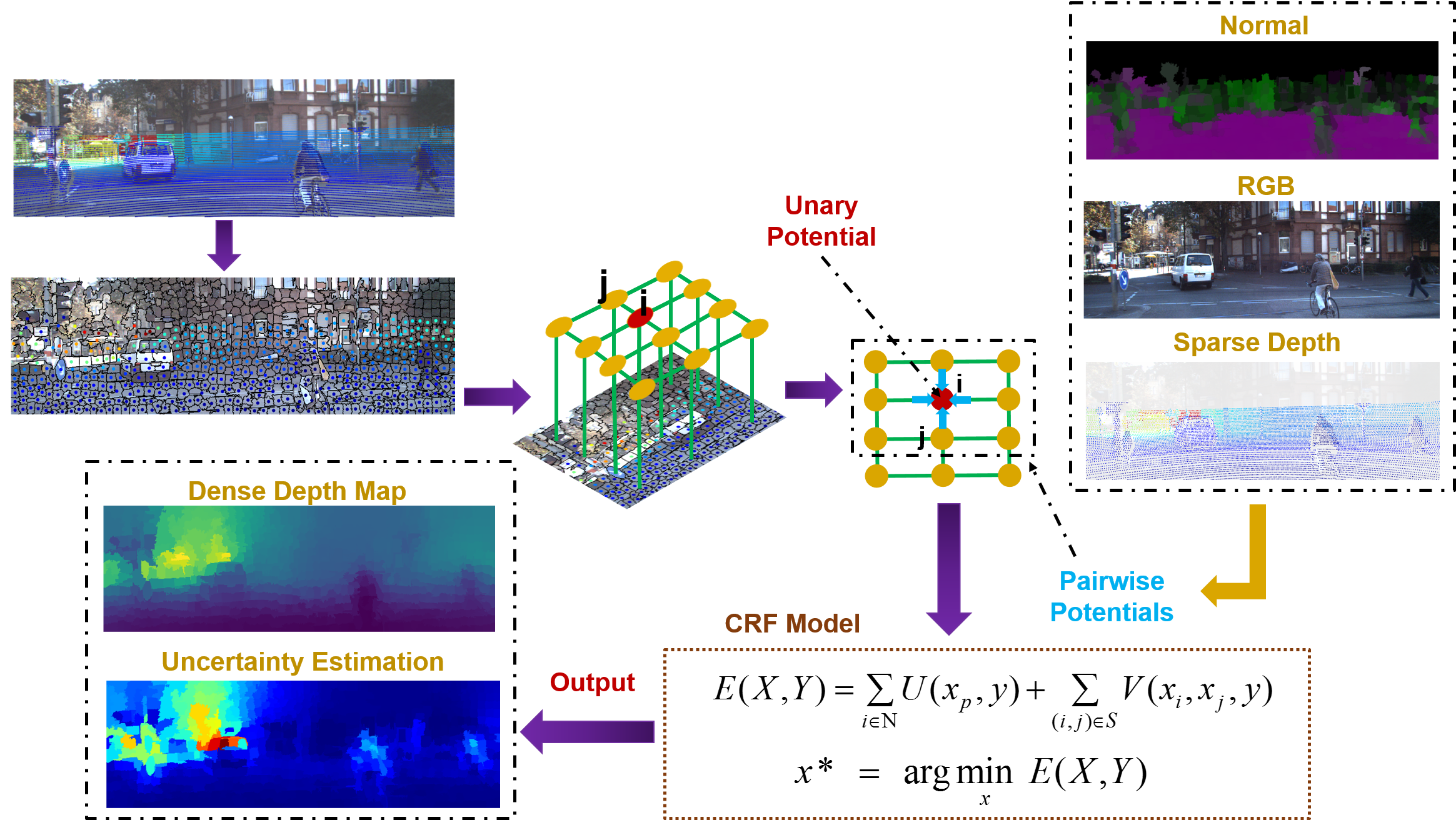}
    \caption{\textbf{Illustration of the proposed model.} On the top left is a fused view of the image and LIDAR point cloud on superpixels.
On the top right are the normal surface map and RGB inputs used in the pairwise potentials.
On the top middle is the graph structure of the CRF: The yellow nodes represent the centroids of the image superpixels and the green branches the connections between them.
    The outputs of the unary part and the pairwise part are then fed to the CRF structured loss layer, which minimizes the corresponding energy function.
On the bottom left is the probabilistic output, a dense depth map and uncertainty estimation map (see text for details).}
    \label{fig:supSRRFmovie}
\end{figure*}

To simplify the solution for the energy function, one can take the negative logarithm of the left hand side and right hand side of the equation of the probability distribution $Pr(X | Y)$: then the problem of maximizing the conditional probability becomes an energy minimization problem.  Therefore, maximizing the probability distribution $\operatorname{Pr}(\mathbf{X} | \mathbf{Y})$ is equivalent to minimizing the corresponding energy function:

$$
\mathbf{x}^{\star}=\arg \min _{\mathbf{x}} E(\mathbf{X}, \mathbf{Y}).
$$

We formulate the energy function as a typical combination of unary potentials $U$ and pairwise potentials $V$ over the nodes (superpixels) $N$ and edges $S$ of the image $x$:

$$
E(\mathbf{X}, \mathbf{Y})=\sum_{p \in \mathcal{N}} U\left(x_{p}, \mathbf{y}\right)+\sum_{(p, q) \in \mathcal{S}} V\left(x_{p}, x_{q}, \mathbf{y}\right)
$$

The unary term $U$ aims to regress the depth value from a single superpixel.
The pairwise term $V$ encourages neighbouring superpixels with similar appearances to take similar depths \cite{A31, A40}.

\subsection{Potential functions}

The proposed multi-modal depth estimation model is composed of unary and pairwise potentials.
For an input image, which has been over-segmented into $n$ superpixels, we define a unary potential for each superpixel.
The pairwise potentials are defined over the four-neighbour vicinity of each superpixel.

The unary potentials are built by aggregating all LiDAR observations inside each superpixel.
The pairwise part is composed of similarity vectors, each with $K$ components, that measure the agreement between different features of neighbouring superpixel pairs.
Therefore, we explicitly model the relations between neighbouring superpixels through pairwise potentials.
In the following, we describe the details of the potentials involved in our energy function.


\subsubsection{ Unary potential}
The unary potential is constructed from the LiDAR sensor measurements by considering the least square loss between the estimated $x_{i}$ and observed $y_{i}$ depth values:

$$ \Phi(\mathbf{x}, \mathbf{y})=\sum_{i \in \mathscr{L}} \sigma_{i}\left(x_{i}-y_{i}\right)^{2}$$

$$
 \Phi (\mathbf{x}, \mathbf{y})=\|\mathbf{W}(\mathbf{x}-\mathbf{y})\|^{2}
$$

\noindent where $\mathscr{L}$ is the set of indices for which a depth measurement is available, and $\sigma_{i}$ is a constant weight placed on the depth measurements.
This potential measures the quadratic distance between the estimated range $X$ and the measured range $Y$, where available.
Finally, in order to write the unary potential in a more efficient matrix form, we define the diagonal matrix $W$ with entries
$$
\mathbf{W}_{i, i}=\left\{\begin{array}{ll}{\sigma_{i}} & {\text { if } i \in \mathscr{L}} \\ {0} & {\text { otherwise }}\end{array}\right.
$$

\subsubsection{Colour pairwise potential}
We construct a pairwise potential from $K$ types of similarity observations, each of which enforces smoothness by exploiting colour consistency features of the neighbouring superpixels.
This pairwise potential can be written as

$$\Psi^{c}(\mathbf{x}, \mathbf{I})=\sum_{i} \sum_{j \in \mathscr{N}(i)} e_{i, j}\left(x_{i}-x_{j}\right)^{2}$$

$$
\Psi^{c}(\mathbf{x}, \mathbf{I})=\|\mathbf{S} \mathbf{x}\|^{2}
$$

\noindent where $I$ is an RGB image, $\mathscr{N}(i)$ is the set of horizontal and vertical neighbours of $i$, and each row of $S$ represents the weighting factors for pairs of adjacent range nodes.
As the edge strength between nodes, we use an exponentiated $L_{2}$ norm of the difference in pixel appearance.

$$
e_{i, j}=\exp -\frac{\left\|\mathbf{c}_{i}-\mathbf{c}_{j}\right\|^{2}}{\sigma_{d}^{2}}
$$

\noindent where $\mathbf{c}_{i}$ is the RGB colour vector of pixel $i$ and $\sigma_{d}$ is a tuning parameter.
A small value of $\sigma_{d}$ increases the sensitivity to changes in the image. Thanks to this potential, the lack of content or features in the RGB image is considered by our model as indicative of a homogeneous depth distribution, in other words, a planar surface.

\subsubsection{ Surface-normal pairwise potential}
The mathematical formulation of this potential is similar to the previous colour potential. However, the surface-normal potential considers surface normal similarities instead of colour.
The weighting factors $nr_{i, j}$ for this case are formulated using the cosine similarity, which is a measure of the similarity between two non-zero vectors of an inner product space that employs the cosine of the angle between them.
The cosine of $0$ is $1$, and it is less than $1$ for any angle in the interval $[0, \pi]$ radians. It is thus a measurement of orientation instead of magnitude \cite{A53}.The cosine of two non-zero vectors can be found by using the Euclidean dot product formula: 

$$\mathbf{A} \cdot \mathbf{B}=\|\mathbf{A}\|\|\mathbf{B}\| \cos \theta$$

Therefore, the cosine similarity can be expressed by

$$ \cos (\theta)=\frac{\mathbf{A} \cdot \mathbf{B}}{\|\mathbf{A}\|\|\mathbf{B}\|} = \frac{\sum_{t=1}^{n} A_{i} B_{i}}{\sqrt{\sum_{t=1}^{n} A_{i}^{2}} \sqrt{\sum_{t=1}^{n} B_{i}^{2}}}$$

\noindent where $A_i$ and $B_i$ are the components of vectors $A$ and $B$, respectively.
Finally, we define our surface normal potential by the following equations.

$$\Psi^{n}(\mathbf{x}, \mathbf{In})=\sum_{i} \sum_{j \in \mathscr{N}(i)} nr_{i, j}\left(x_{i}-x_{j}\right)^{2}$$

$$
\Psi^{n}(\mathbf{x}, \mathbf{In})=\|\mathbf{P} \mathbf{x}\|^{2} 
$$

$$ nr_{i, j} =\frac{\sum_{t=1}^{n} In_{i}  In_{j}}{\sqrt{\sum_{t=1}^{n} In_{i}^{2}} \sqrt{\sum_{t=1}^{n} In_{j}^{2}}}$$

\subsubsection{Depth pairwise potential}
This pairwise potential encodes a smoothness prior over depth estimates which encourages neighbouring superpixels in the image to have similar depths.
Usually, pairwise potentials are only related to the colour difference between pairs of superpixels. However, depth smoothness is a valid hypothesis which can potentially enhance depth inference.

To enforce depth smoothness, a distance-aware Potts model was adopted.
Neighbouring points with smaller distances are considered to be more likely to have the same depth.

The mathematical formulation of this potential is similar to the colour pairwise potential, as it follows the Potts model:

$$\Psi^{d}(\mathbf{x}, \mathbf{D})=\sum_{i} \sum_{j \in \mathscr{N}(i)} e_{i, j}\left(x_{i}-x_{j}\right)^{2}$$

\noindent and the weighting factor $dp_{i, j}$ for this case is formulated as

$$
dp_{i, j}=\exp -\frac{\left\|\mathbf{p}_{i}-\mathbf{p}_{j}\right\|^{2}}{\sigma_{p}^{2}}
$$

\noindent where $\mathbf{p}_{i}$ is the 3D location vector of the LiDAR point $i$ and $\sigma_{p}$  is a parameter controlling the strength of enforcing close points to have similar depth values.

\subsubsection{Uncertainty potential:}
Depth uncertainty estimation is important for refining depth estimation  \cite{A56, A58}, and in safety critical systems \cite{A57}. It allows an agent to identify unknowns in an environment in order to reach optimal decisions.
Our method provides uncertainties for the estimates of the pixel-wise depths by taking into account the number of LiDAR points present for each superpixel. The uncertainty potential is similar to the unary potential. It is constructed from the number of LiDAR points projected onto a superpixel, and employs the following least square loss:

$$ U^{c}(\mathbf{x}, \mathbf{y})=\sum_{i \in \mathscr{L}} \sigma_{i}\left(x_{i}-unc_{i}\right)^{2}$$

$$
 U^{c}(\mathbf{x}, \mathbf{y})=\|\mathbf{W}(\mathbf{x}-\mathbf{unc})\|^{2}
$$

\noindent where $\mathbf{unc}$ is defined as follows:

$$
\mathbf{unc}_{i, i}=\left\{\begin{array}{ll}{\sigma_{i}} & {\text { if P projected on SPx is 0}}\\
{\psi {i}}& {\text { if P projected on SPx is >0 and <2}}
\\
{mean} & {\text { otherwise}}\end{array}\right.
$$

\noindent where $P$ is a 3D point and SPx is a superpixel. In locations with accurate and sufficiently many LiDAR points, the model will produce depth predictions with a high confidence.
This uncertainty estimation provides a measure of how confident the model is about the depth estimation.
This results in an overall better performance, since uncertain estimates with high uncertainty can be neglected by higher level tasks that use the estimated depth maps as an input.

\subsection{Optimization} 

With the unary and the pairwise potentials defined, we can now write the energy function as

$$ E(\mathbf{X}, \mathbf{Y}) = \left(\alpha \right)\Phi(\mathbf{x}, \mathbf{y})+(\beta )\Psi^{c}(\mathbf{x}, \mathbf{I}) \:.\:.\:.
$$

\begin{equation}
\label{eq2}
\qquad+\:.\:.\:.\: (\gamma)\Psi^{n}(\mathbf{x}, \mathbf{In})+(\delta)\Psi^{d}(\mathbf{x}, \mathbf{In})
\end{equation}

The scalars $\alpha$, $\beta$, $\gamma$, $\delta$ $\in$ [0,1] are weightings for the four terms.
We may further expand the unary and pairwise potentials to

\begin{equation}
\label{eq3}
\Phi(\mathbf{x}, \mathbf{y})= \alpha(\mathbf{x}^{\mathrm{T}} \mathbf{W}^{\mathrm{T}} \mathbf{W} \mathbf{x}-2 \mathbf{z}^{\mathrm{T}} \mathbf{W}^{\mathrm{T}} \mathbf{W} \mathbf{x}+\mathbf{z}^{\mathrm{T}} \mathbf{W}^{\mathrm{T}} \mathbf{W} \mathbf{z})
\end{equation}

\begin{equation}
\label{eq4}
\enskip\qquad\qquad\Psi^{c}(\mathbf{x}, \mathbf{In}) = \beta(\mathbf{x}^{\mathbf{T}} \mathbf{S}^{\mathbf{T}} \mathbf{S}_{\mathbf{X}})
\end{equation}

\begin{equation}
\label{eq5}
\enskip\qquad\qquad\Psi^{n}(\mathbf{x}, \mathbf{In})=\gamma(\mathbf{x}^{\mathbf{T}} \mathbf{P}^{\mathbf{T}} \mathbf{P}_{\mathbf{X}})
\end{equation}

\begin{equation}
\label{eq6}
\enskip\qquad\qquad\Psi^{d}(\mathbf{x}, \mathbf{In})=\delta(\mathbf{x}^{\mathbf{T}} \mathbf{D}^{\mathbf{T}} \mathbf{D}_{\mathbf{X}})
\end{equation}

We shall pose the problem as one of finding the optimal range vector $\mathbf{x}^{*}$ such that:

$$
\mathbf{x}^{*}=\underset{\mathbf{x}}
{\operatorname{argmin}}\left\{E(\mathbf{X}, \mathbf{Y}) \right\}
$$



Substituting equations \ref{eq3}, \ref{eq4}, \ref{eq5} and \ref{eq6} into \ref{eq2} and solving for $x$ reduces the problem to: $ \mathbf{A x}=\mathbf{b}$ where 

$$
\mathbf{A}= \alpha(\mathbf{W}^{\mathbf{T}}\mathbf{W})+ \beta(\mathbf{S}^{\mathbf{T}}\mathbf{S}) +  \gamma(\mathbf{P}^{\mathbf{T}}\mathbf{P}) + \delta(\mathbf{D}^{\mathbf{T}}\mathbf{D})
$$

$$
\mathbf{b}= \alpha(\mathbf{W}^{\mathbf{T}} \mathbf{W} \mathbf{z})
$$

All we need to do to perform the optimization is to solve a large sparse linear system.
The methods for solving sparse systems are divided into two categories: direct and iterative.
Direct methods are robust but require large amounts of memory as the size of the problem grows.
On the other hand, iterative methods provide better performance but may exhibit numerical problems \cite{A54, A55}.
In the present paper, the fast algorithm Conjugate Gradient Squared proposed by Hestenes and Stiefel \cite{A59, A60}, is employed to solve the energy minimization problem.

\subsection{Pseudo-code} 

Algorithm~\ref{alg:the_alg1} provides the complete pseudo-code for our proposed framework, which has been previously illustrated in Figure~\ref{fig:supSRRFmovie}.
In this algorithm, lines 1 to 5 perform the preprocessing, which includes gathering the multi-modal raw data, building a connection graph between pairs of adjacent superpixels, and projecting the clustered LiDAR points onto the image space.

Lines 6 to 11 constitute the core of the approach.  They include constructing the cost function using different potentials (unary and pairwise) to obtain the complete CRF for the depth estimation.
The objective of the pairwise potentials is to smooth the depth regressed from the unary part based on the  neighbouring  superpixels.
The  pairwise  potential  functions are  based  on  standard  CRF  vertex  and  edge  feature  functions studied  extensively  in  \cite{A66}  and  other  papers.
Our model uses both the content information of the superpixels and relation information between them to infer the depth.
\\

\begin{algorithm}
\DontPrintSemicolon
  
  \KwInput{RGB Image and Sparse 3D Point Cloud}
  \KwOutput{Dense Point Cloud}
  Compute point cloud normal vectors
  
  Perform superpixel segmentation
  
  Build a set of edges and nodes considering the definition of the 4-neighbourhood.
  
  z←Project clustered PCL onto image plane
  
  Initialize uncertainty depth map for the whole image
  
  \For{each superpixel (node)}    
        { 
        Calculate unary potential
        
        Calculate Colour pairwise potential
        
        Calculate Surface-normal pairwise potential
        
        Calculate Depth pairwise potential
        
        Calculate Uncertainty potential
        }
  Infer 2D dense depth map

\caption{UAOFusion Network: A multimodal CRF-based method for Camera-LiDAR depth estimation}
\label{alg:the_alg1}
\end{algorithm}

\section{Results and discussion}
\label{RD}
We evaluate our approach on the raw sequences of the KITTI benchmark, which is a popular dataset for single image depth map prediction.
The sequences contain stereo imagery taken from a car driving in an urban scenario.
 The dataset also provides 3D laser measurements from a Velodyne laser scanner, which we use as ground-truth measurements (projected into the stereo images using the given intrinsics and extrinsics in KITTI).
This dataset has been used to train and evaluate the state-of-the-art methods and allows quantitative comparisons.

First, we  evaluate  the prediction accuracy  of  our  proposed method  with different  potentials in Section \ref{sec:arc_ev}.
Second, in Section \ref{sec:num_pix} we  explore  the  impact on the depth estimation  of the number of  sparse  depth  samples and the number of superpixels. 
Third, Section \ref{sec:alg_ev} compares our approach  to state-of-the-art  methods on the KITTI dataset.
Lastly,  in Sections  \ref{sec:app_kitti} and \ref{sec:app_UAO}, we  demonstrate  two  use cases of our proposed algorithm, one for creating LiDAR super-resolution from sensor data provided by the KITTI dataset and another one for a dataset collected in the context of this work.

\subsection{Evaluation Metrics}

We evaluate the accuracy of our method in depth prediction using the 3D laser ground truth on the test images.
We use the following depth evaluation metrics: root mean squared error (RMSE), mean absolute error (MAE) and mean absolute relative error (REL), among which RMSE is the most important indicator and chosen to rank submissions on the leader-board since it measures error directly on depth and penalizes on further distance where depth measurement is more challenging.
These metrics were used by \cite{A11, A28, A62, A64} to estimate the accuracy of monocular depth prediction.\\

$$ RMSE = \sqrt{\frac{1}{|T|} \sum_{d \in T}\|\hat{d}-d\|^{2}}$$

$$MAE = \frac{1}{T} \sum_{d \in T}\|\hat{d}-d\|^{2}$$


$$REL=\frac{1}{T} \sum_{d \in T}\left(\frac{\left|\hat{d}-d\right|}{\hat{d}}\right)$$

\noindent Here, ${d}$ is the ground truth depth, $\hat{d}$ is the estimated depth, and $T$ denotes the set of all points in the test set images.
In order to compare our results with those of Eigen et al. \cite{A11} and Godard et al. \cite{A27}, we crop our image to the evaluation crop applied by Eigen et al. We also use the same resolution of the ground truth depth image and cap the predicted depth at 80 m \cite{A27}.

\subsection{Architecture Evaluation}
\label{sec:arc_ev}
This section presents an empirical study of the impact on the accuracy of the depth prediction of different choices for the potential functions and hyparameters.
In the first experiment, we compare the impact of sequentially adding our proposed pairwise potentials.
We first evaluate a model with only unary and colour pairwise potentials.
Then we added the surface-normal pairwise potential, and finally the depth pairwise potential is included.
As shown in Table 1, the RMSE is improved after adding each pairwise potential.

\begin{table}[width=.7\linewidth,cols=3,pos=h]
\label{table:comp}
\caption{Depth completion errors [mm] after adding pairwise potentials  (lower is better)}\label{tbl1}

\begin{tabular*}{\tblwidth}{@{} LLLL@{} }
\toprule
Algorithm & Potential functions & RMSE \\
\midrule
\textbf{Ours} & I & 865.31\\
\textbf{Ours} & II & 854.24\\
\textbf{Ours} & III  & 849.39\\
\bottomrule
\end{tabular*}
\end{table}

\begin{figure}
	\centering
		\includegraphics[trim={0.8cm 0 0 0},clip,scale=.236]{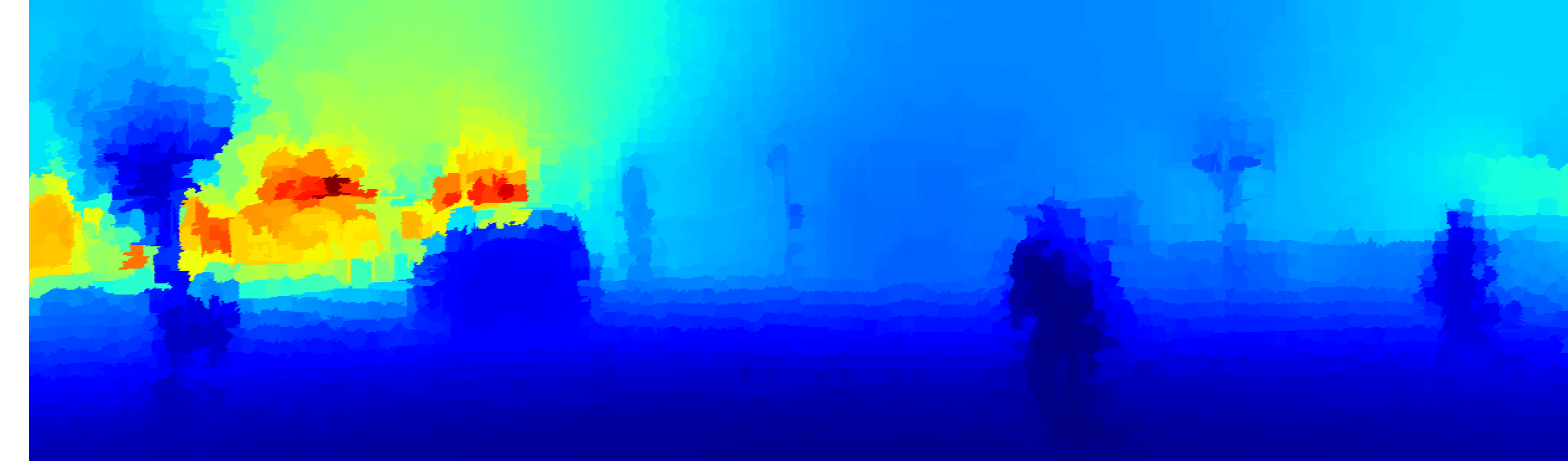}
		\includegraphics[trim={0.8cm 0 0 0},clip,scale=.236]{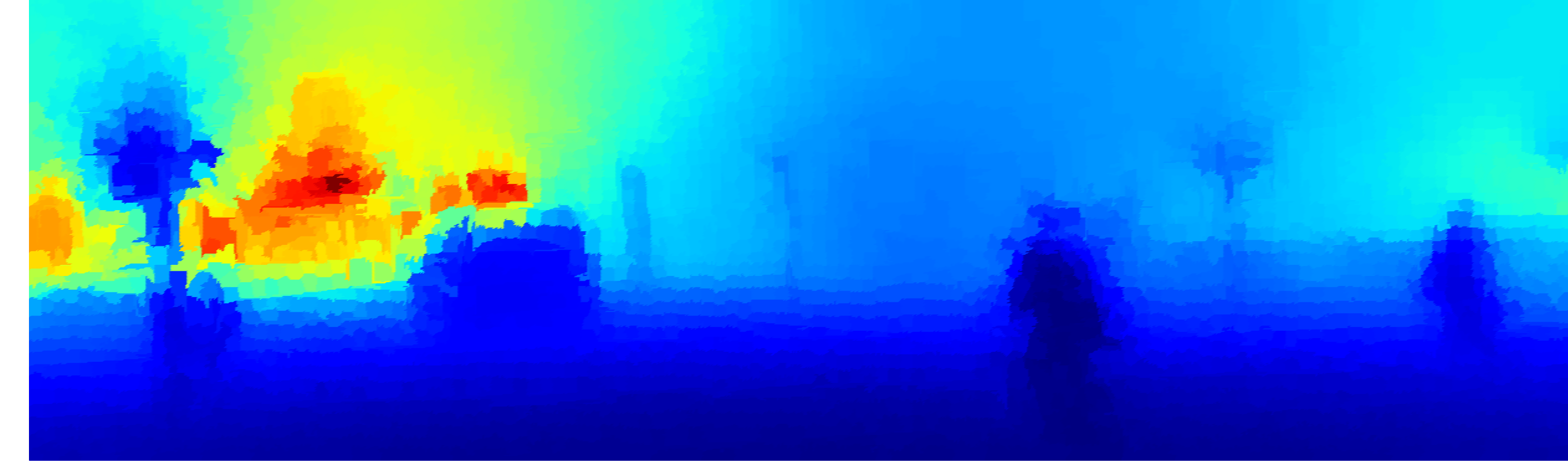}
		
		\includegraphics[trim={0.8cm 0 0 0},clip,scale=.236]{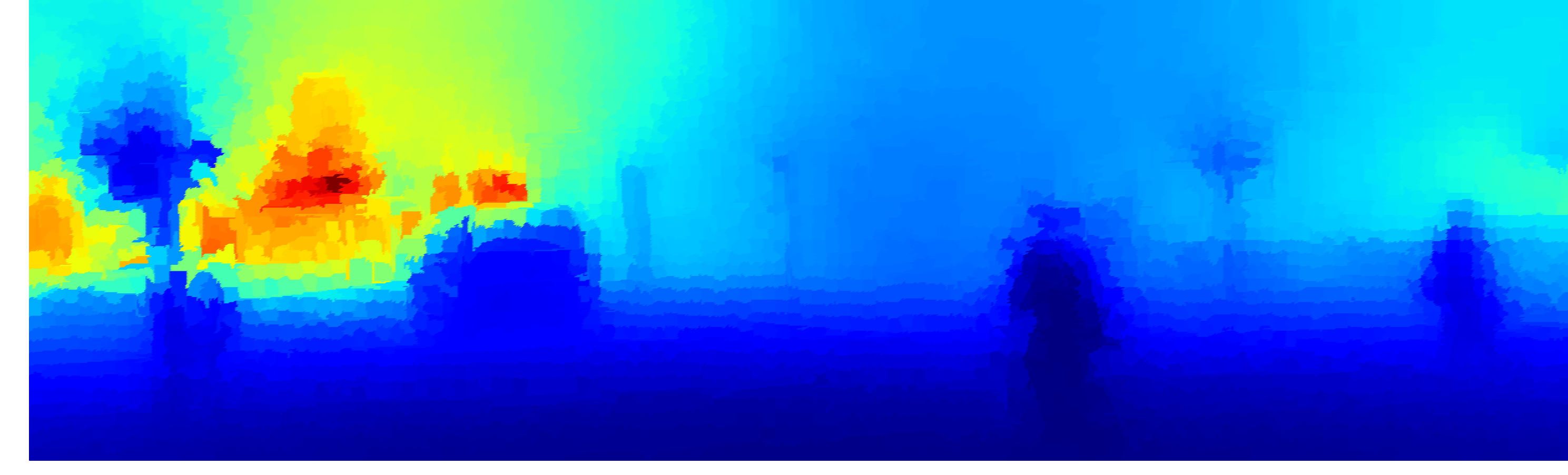}
		
	\caption{Qualitative  evaluation  of  the  impact  of  the pairwise potentials defined as CRF terms.
	In  row  order: 1st: Pairwise potential I, penalizes dissimilar depth estimates of neighbouring pixels which have similar colours in the RGB image, 2nd:  pairwise potential II, penalizes the depth differences between neighbouring superpixels whose normal surface vectors have large cosine similarities, 3rd: pairwise potential III, penalizes neighbouring superpixels with large observed depth differences.}
		
	\label{FIG:5}
\end{figure}


\subsection{The number of superpixels}
\label{sec:num_pix}
In this section, we explore the relation between the prediction accuracy and the number of available depth samples and the number of superpixels.

\begin{figure}
	\centering
        \includegraphics[trim={0.8cm 0 0 180},clip,scale=.304]{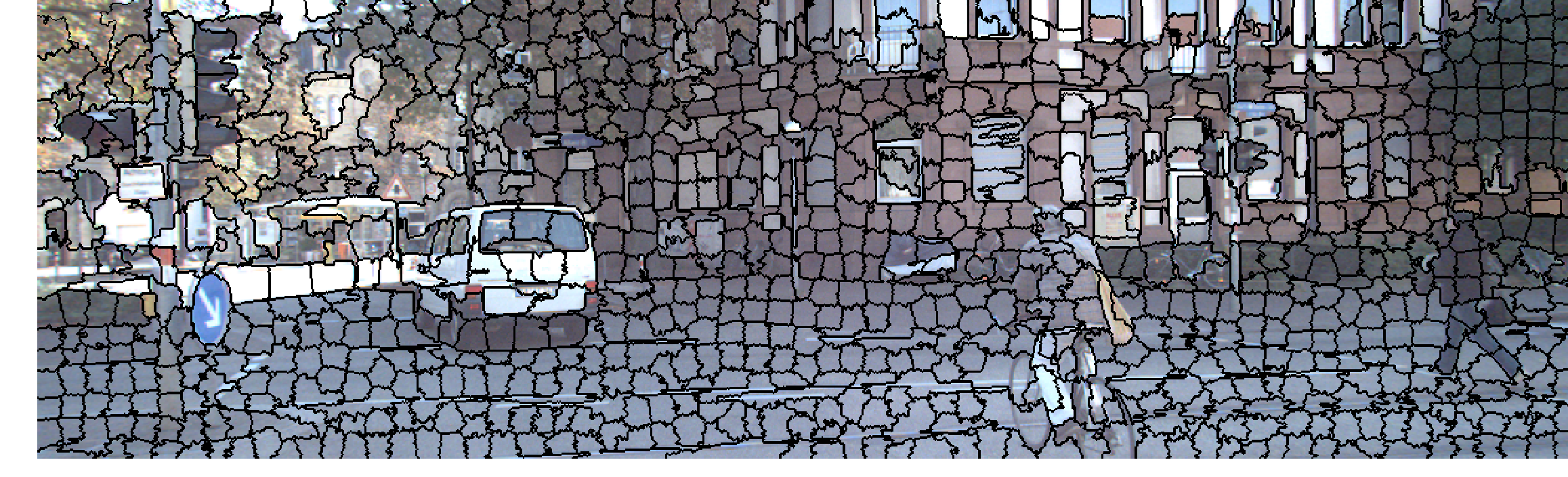}

		\includegraphics[trim={0.8cm 0 0 0},clip,scale=.236]{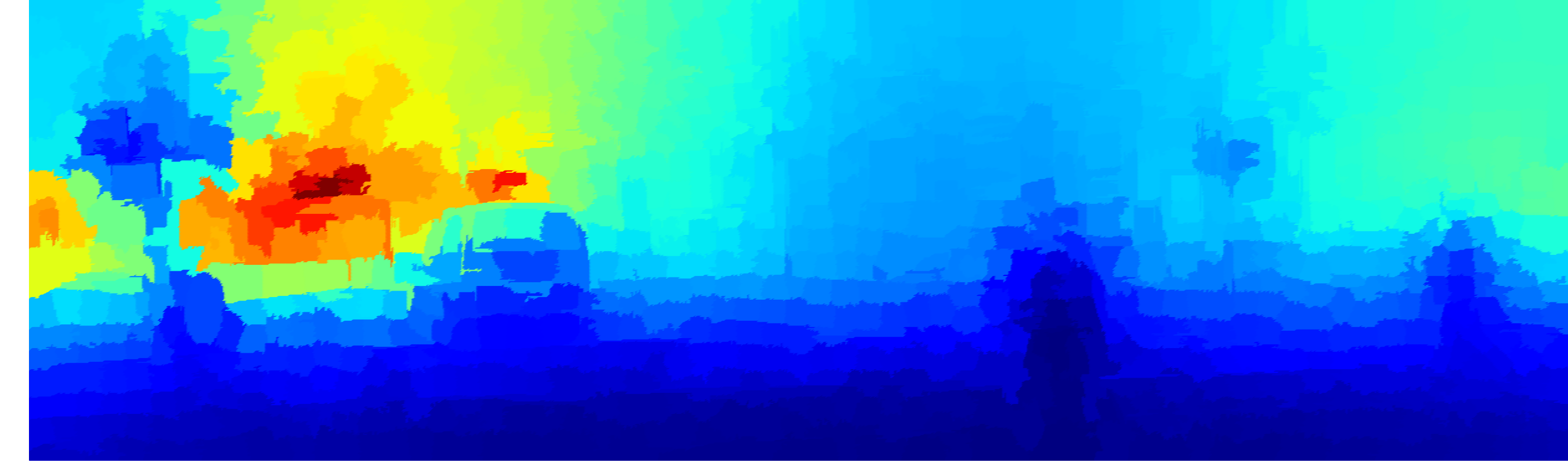}

        \includegraphics[trim={0.8cm 0 0 180},clip,scale=.304]{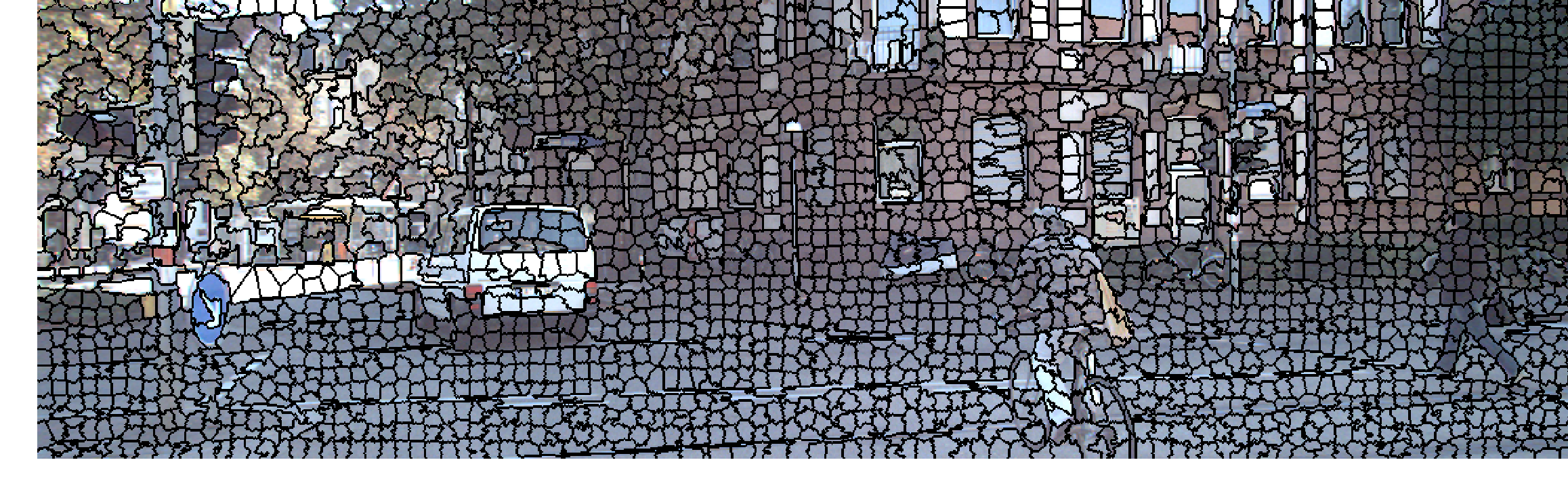}
		\includegraphics[trim={0.8cm 0 0 0},clip,scale=.236]{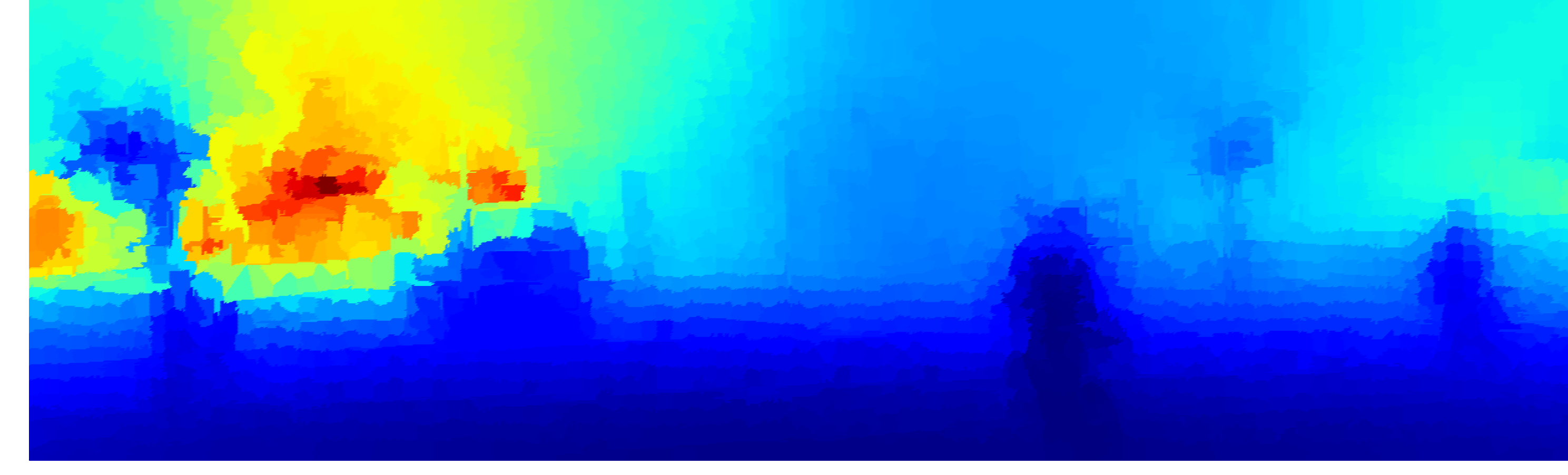}
		
        \includegraphics[trim={0.8cm 0 0 180},clip,scale=.304]{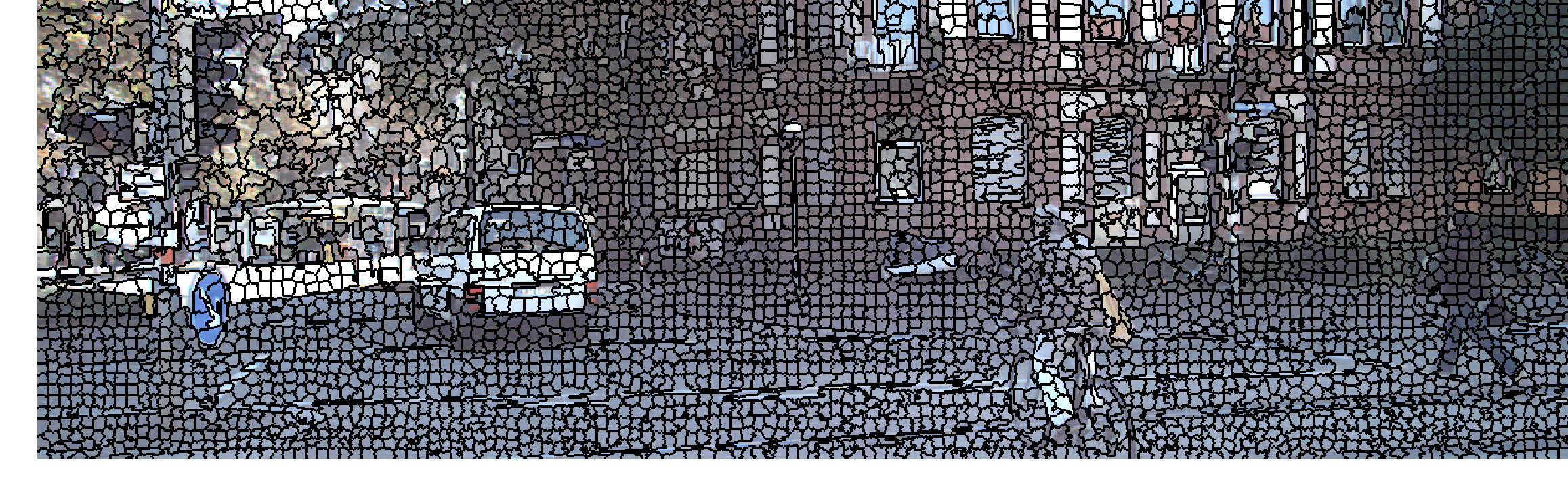}
		\includegraphics[trim={0.8cm 0 0 0},clip,scale=.236]{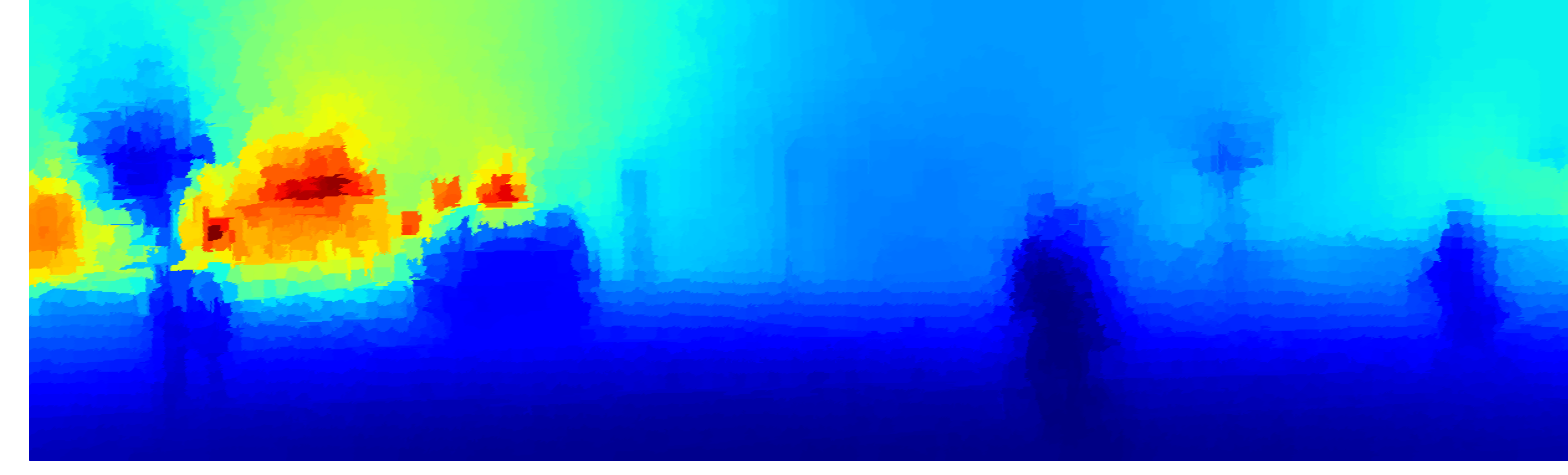}
	\caption{Visual comparison  of dense depth maps produced by the CRF framework when varying the size of the superpixels.
From top to bottom, 1200, 2400 and 5500 superpixels.
}
		
	\label{FIG:6}
\end{figure}

\begin{figure}
	\centering
        \includegraphics[width=0.4\textwidth]{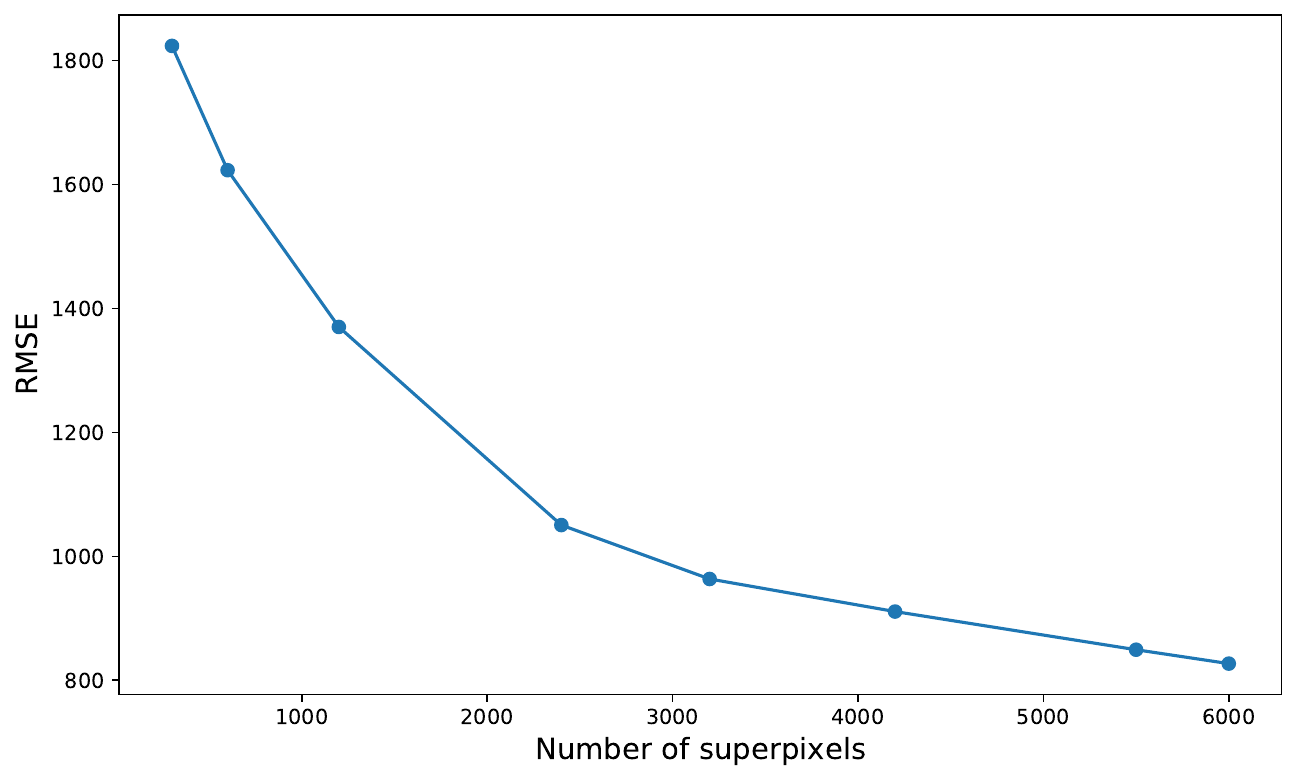}
	\caption{Convergence for different number of  superpixels.
}
		
	\label{FIG:convergence}
\end{figure}

As displayed in \textcolor{blue}{Fig. 5}, a greater number of super pixels yields better results in error measurements. Although a larger number of sparse depth observations improves the quality of the depth map, the performance converges when the number of superpixels is more than 5000, which is about 1.5\% of the total number of pixels. We ran an exhaustive evaluation of our method for a different amount of superpixels. \textcolor{blue}{Fig. \ref{FIG:convergence}} clearly shows that our method's error decreases with an increased number of superpixels.

\begin{table}[width=.7\linewidth,cols=3,pos=h]
\caption{Depth completion errors [mm] for different number of superpixels (lower is better)}\label{tbl1}
\begin{tabular*}{\tblwidth}{@{} LLLL@{} }
\toprule
Algorithm & \#Superpixels & RMSE\\
\midrule
\textbf{Ours} & 1200 & 1370.27  \\
\textbf{Ours} & 2400 & 1050.55  \\
\textbf{Ours} & 5500 & 849.39 \\
\bottomrule
\end{tabular*}
\end{table}

\subsection{Sub-sampling 3D depth points}
\label{sec:depth}
\begin{figure}
	\centering
        \includegraphics[width=0.4\textwidth]{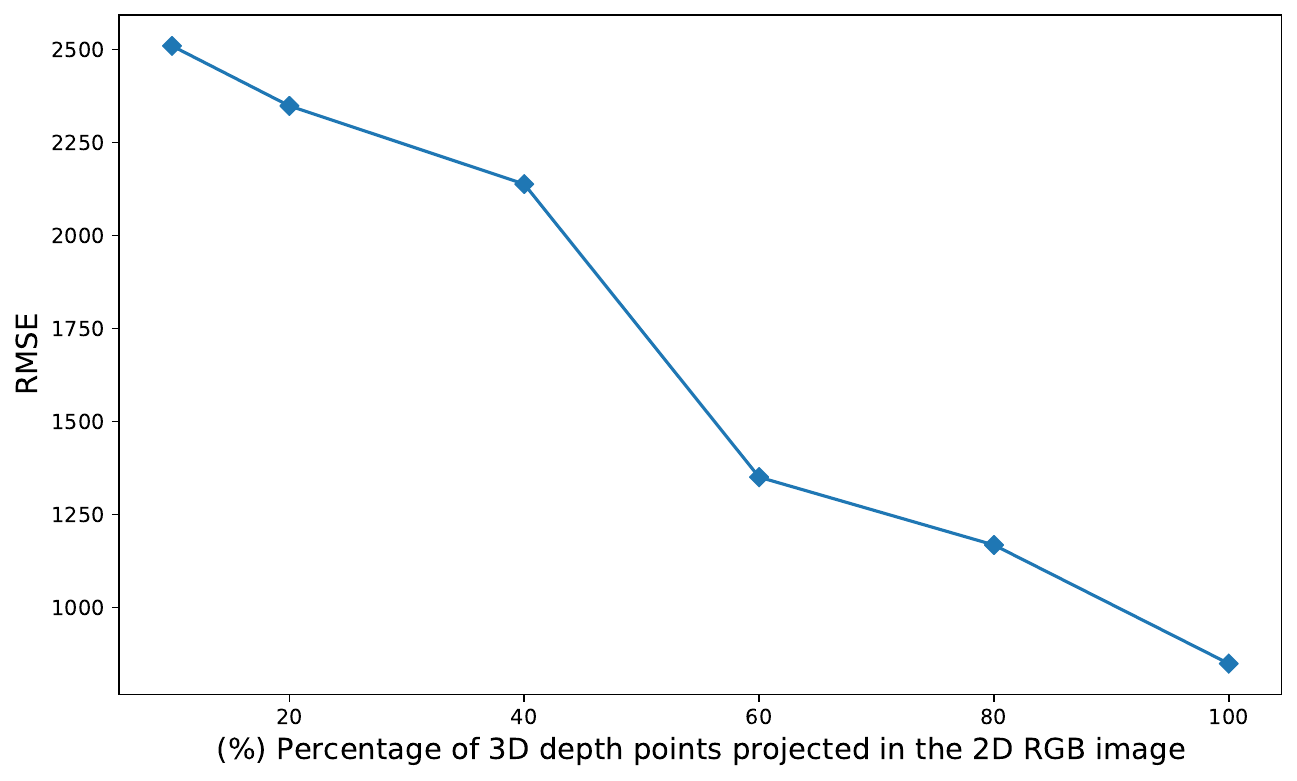}
	\caption{Convergence for different number of 3D depth points projected into 2D RGB images. The amount of 3D depth points used is represented as a percentage ($\%$). 100\% means we are using all the 3D depth points projected into the 2D image.}		
	\label{FIG:depth_subsampling}
\end{figure}

We performed a quantitative analysis of the impact of observed 3D point sparsity on the error of our proposed method, by decreasing the amount of 3D points considered during inference. As it is shown in \textcolor{blue}{Fig.~\ref{FIG:depth_subsampling}}, our method's error decreases with an increased number of 3D depth points, enabling a better dense depth map estimation. From \textcolor{blue}{Fig.~\ref{FIG:depth_subsampling}}, we can argue that the amount of 3D depth points is really important for an accurate depth map estimation. Even though our estimation error increases with the sparsity of the depth observations, our method manages to provide state-of-the-art performances even when the depth observations are sampled down to 40\%.

\subsection{Algorithm Evaluation for Depth Completion}
\label{sec:alg_ev}

This is a more challenging dataset than other datasets for depth estimation:
the distances in the KITTI dataset are larger than in other datasets, e.g. NYU-Depth-V2 dataset.  Hence, the KITTI odometry dataset is more challenging for the depth estimation task.

The performance of our method and those of other existing methods on the KITTI dataset are shown in Table 3. Table 3 shows that the proposed method outperforms other depth map estimation approaches which are well-accepted in the robotics community. Our model relies on the number of superpixels and the resolution of input data sources. This means that the model's performance will increase if we increase the number of superpixels, the image resolution, and the density of the LiDAR data.

\begin{table}[width=\linewidth,cols=4,pos=h]
\caption{Depth completion errors [mm] by different methods on the test set of KITTI depth completion benchmark (lower is better)}\label{tbl1}
\begin{tabular*}{\tblwidth}{@{} LLLL@{} }
\toprule
Algorithm & RMSE & MAE\\
\midrule

Semantically guided &  &  \\
depth upsampling \cite{A78} & 2312.57 & 605.47 \\
Convolutional spatial &  &  \\
propagation network (CSPN) \cite{A75} & 1019.64 & 279.46  \\
Hierarchical multi-scale &  &  \\
sparsity-invariant network (HMS-Net) \cite{A79} & 841.77 & 253.47 \\
End-to-end sparse-to-dense &  &   \\
network (S2DNet) \cite{A77} & 830.57 & 247.85  \\
Self-supervised &  &  \\
sparse-to-dense network \cite{A36} & 814.73 & 249.95 \\
\textbf{Ours} & 849.39 & 263.31 \\  
\bottomrule
\end{tabular*}
\end{table}

\begin{figure}
	\centering
		\includegraphics[trim={0.8cm 0 0 0},clip,scale=.235]{figs/OK1.png}
		\includegraphics[trim={0.8cm 0 0 0},clip,scale=.30]{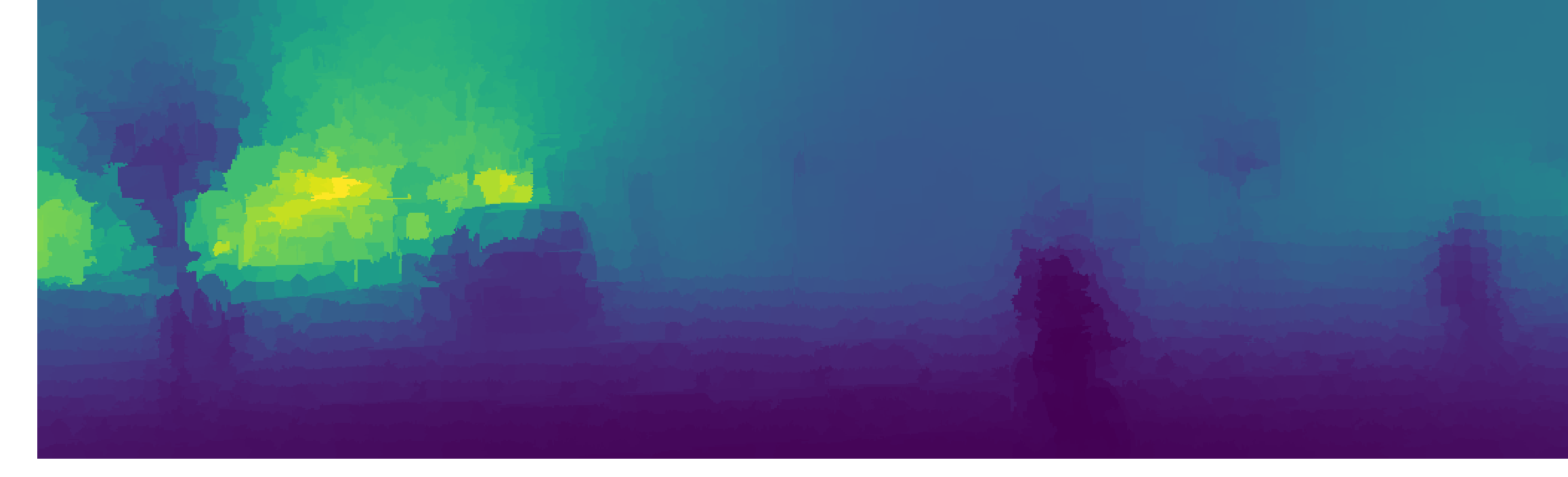}
		\includegraphics[trim={0.8cm 0 0 0},clip,scale=.235]{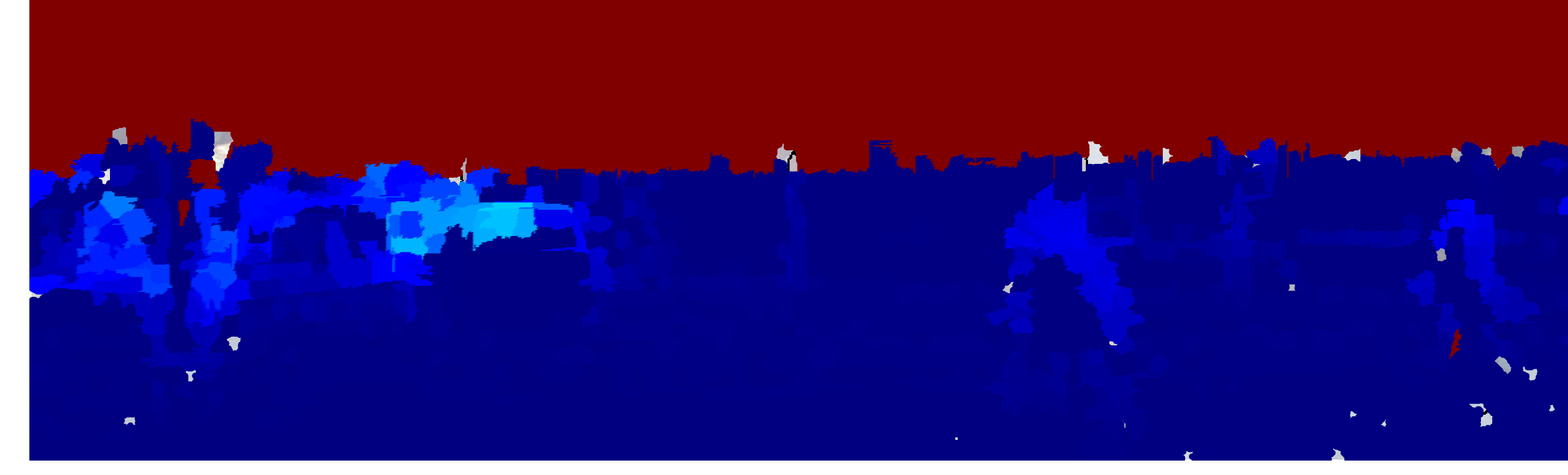}
		\includegraphics[trim={0.8cm 0 0 0},clip,scale=.235]{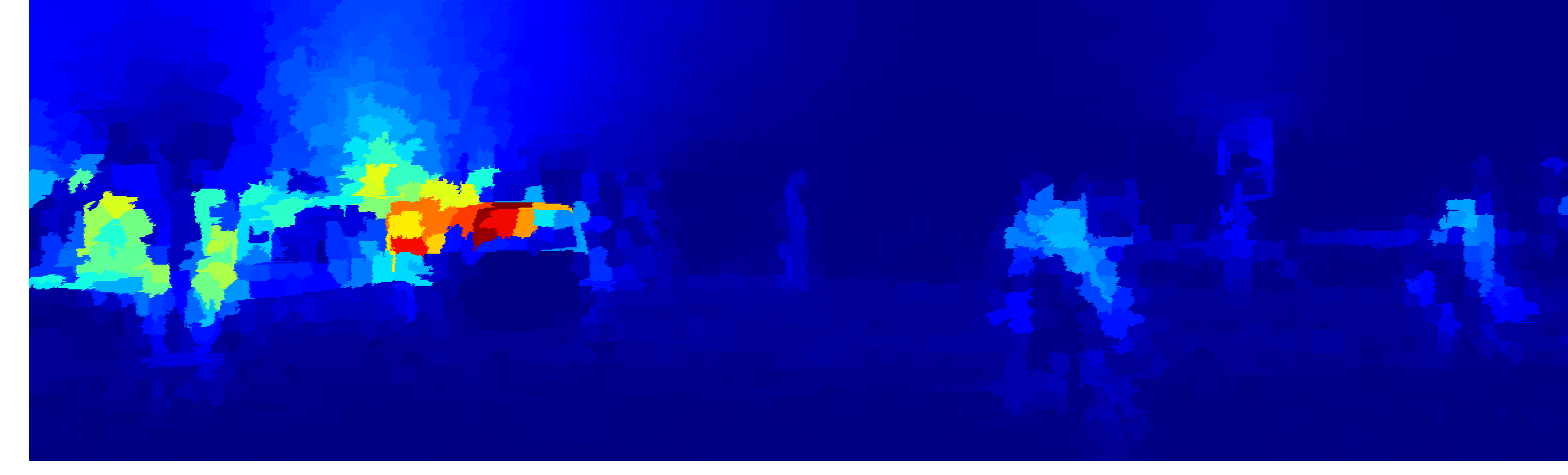}
		
	\caption{Depth completion and uncertainty estimates of our approach on the KITTI raw test set.
From top to bottom: RGB and raw depth projected onto the image; high-resolution depth map; raw uncertainty; and estimated uncertainty map.}
		
	\label{FIG:7}
\end{figure}

\subsection{Algorithm Evaluation for LiDAR Super-Resolution}
\label{sec:app_kitti}

We present another demonstration of our method in super-resolution of LiDAR measurements.
3D LiDARs have a low vertical angular resolution and thus generate a vertically sparse point cloud.
We use all measurements in the sparse depth image and RGB images as input to our framework.
An example is shown in \textcolor{blue}{Fig.~4}.
The cars are much more recognizable in the prediction than in the raw scans.

\begin{figure}
	\centering
		\includegraphics[trim={0.8cm 0 0 0},clip,scale=.30]{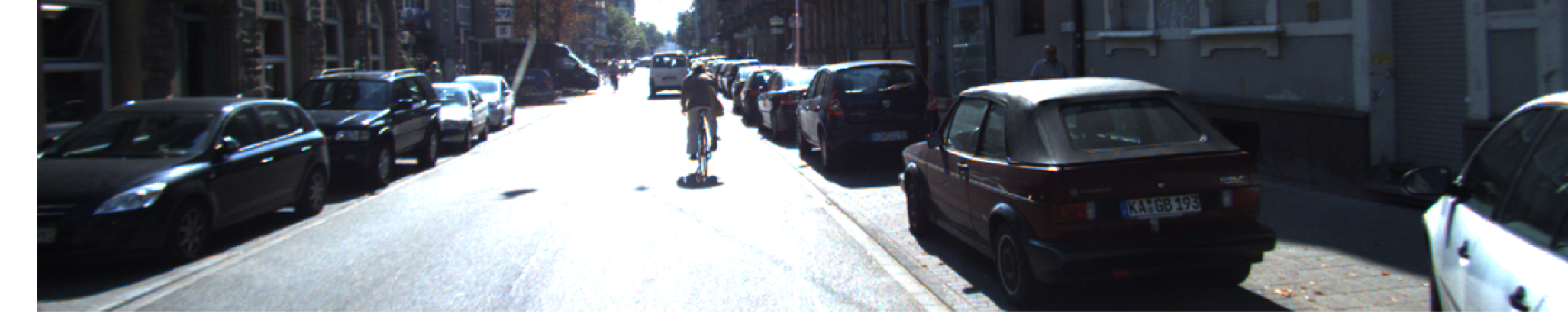}

		\includegraphics[trim={0.8cm 0 0 0},clip,scale=.236]{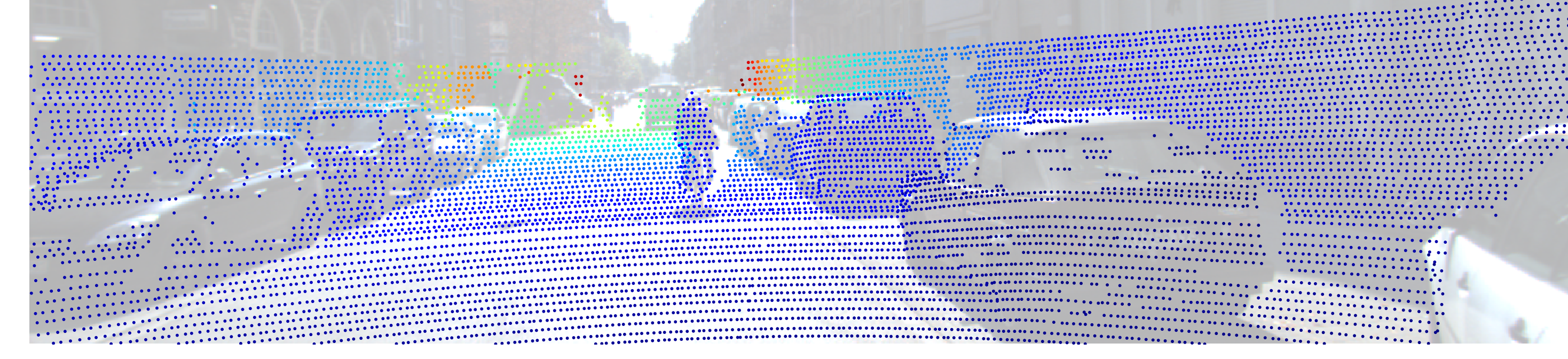}	
		
		\includegraphics[trim={0.8cm 0 0 0},clip,scale=.30]{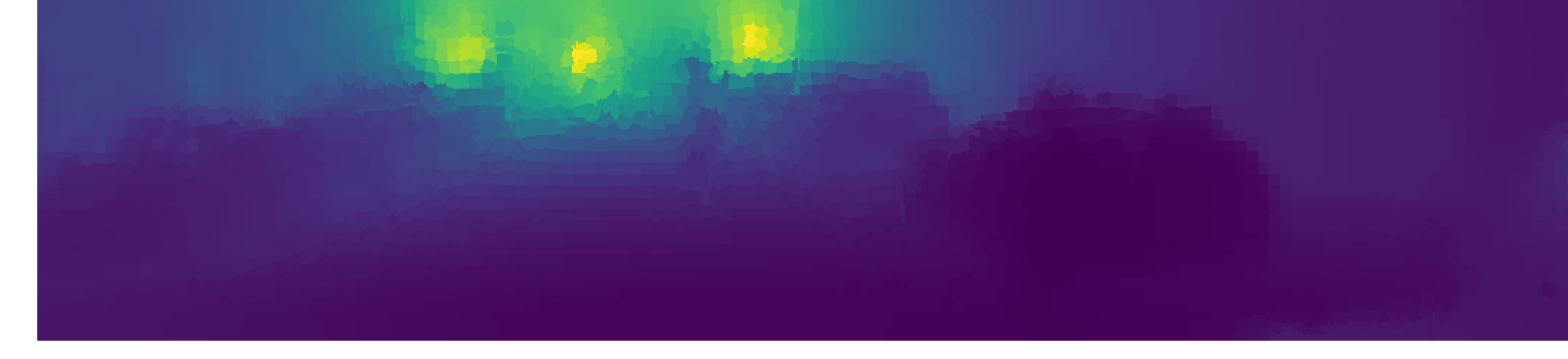}

		\includegraphics[trim={0.8cm 0 0 0},clip,scale=.301]{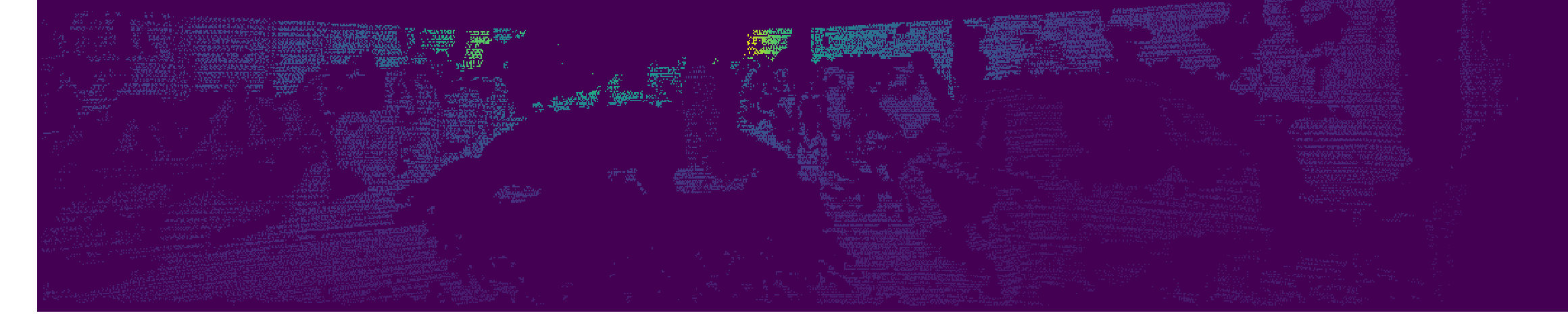}

	\caption{LiDAR super-resolution.
Creating dense point clouds from sparse raw measurements. From top to bottom: RGB image, raw depth map, predicted depth and ground truth depth map. Distant cars are almost invisible in the raw depth map, but are easily recognizable in the predicted depth map}
	\label{FIG:8}
\end{figure}

On the other hand, starting from a LiDAR Super-Resolution map we can generate a 3D reconstruction of the scene. The reconstruction of three-dimensional (3D) scenes has many important applications, such as autonomous navigation \cite{A67}, environmental monitoring \cite{A68} and other computer vision tasks \cite{A69, toro2019divide}.

Therefore, a dense and accurate model of the environment is crucial for autonomous vehicles.
In fact, imprecise representations of the vehicle’s surrounding may lead to unexpected situations that could endanger the passengers.
In this paper, the 3D modeling is generated using a combination of image and range data is a sensor fusion approach that takes the strengths of each in order to overcome their limitations.
Images normally have higher resolution and more visual information than range data, and range data are noisy, sparse, and have less visual information, but already contain 3D information.

The qualitative and quantitative results presented here suggest that our system provides 3D reconstructions of reasonable quality. Following \cite{A29}, we use a random subset of 2000 images from the test sequences for evaluation. 
We take the bottom part 912×228 due to there being no depth at the top area, and only evaluate the pixels with ground truth. The performance of our approach and state-of-the-art depth completion methods are presented in Table 4.

\begin{table}[width=1\linewidth,cols=4,pos=h]
\caption{Depth estimation errors [m] by different methods on the test set of KITTI depth estimation benchmark (lower is better)}\label{tbl1}
\begin{tabular*}{\tblwidth}{@{} LLLL@{} }
\toprule
Algorithm & RMSE[m] & REL & Log10\\
\midrule
Multi-modal Auto-Encoders \cite{A74} & 7.14 & 0.179 & - \\
Residual of residual network \cite{A73} & 4.51 & 0.113 & 0.049 \\
Residual Up-Projection \cite{A76} & 3.67 & 0.072 & - \\
Sparse-to-dense \cite{A29} & 3.37 & 0.073 & - \\
CSPN \cite{A75} & 3.24 & 0.059 & - \\
S2DNet \cite{A77} & 3.11 & 0.069 & 0.038 \\

\textbf{Ours} & 3.59 & 0.072 & 0.041 \\
\bottomrule
\end{tabular*}
\end{table}

In Table 4, the RMSE value of Sparse-to-dense, CSPN and S2DNet methods is slightly better (lower) than our approach. All these three methods solve the depth estimation task through deep learning algorithms which require large amounts of data, which can be very costly in practice, and additional techniques like data augmentation to improve their performance.  For example, S2DNet uses a dataset composed of 1,070,568 images, while our approach only used 42\% of that amount of data plus the LiDAR information. Additionally, these deep learning methods make use of advanced computing resources which limit their use in real-world applications. 

Our method is very competitive with deep learning models as is shown in Table 4, without demanding a large amount of data or additionally strategies. Moreover, we highlight the fact that our method's evaluation does not consider the full image data. As displayed in \textcolor{blue}{Fig. \ref{FIG:convergence}}, our method converges to a better solution when we use 6000 superpixels instead of 5500. We do not evaluate further because the trend is clear, as much superpixels we use, a better dense depth map is predicted.

\subsection{Depth estimation with very sparse LiDAR data: The UAO LiDAR-RGB Dataset}
\label{sec:app_UAO}

Thus far, we have sampled the depth from high-quality LiDAR depth maps, but in practice, sparse depth inputs may come from less reliable sources. Therefore we provide a qualitative evaluation of this model on our own well-calibrated LiDAR and  RGB  dataset. We  use a 16-beam LiDAR  along  with  a  Stereo Labs Zed Mini camera with 1280×720 resolution. This dataset enables us to demonstrate the stability and robustness of the proposed model in particularly challenging scenarios. The scenes were recorded with a low resolution of the camera and the LiDAR sensor in comparison to the KITTI benchmark.

\begin{figure}
	\centering
		
		\includegraphics[trim={0.8cm 0 0.3cm 0},clip,scale=.238]{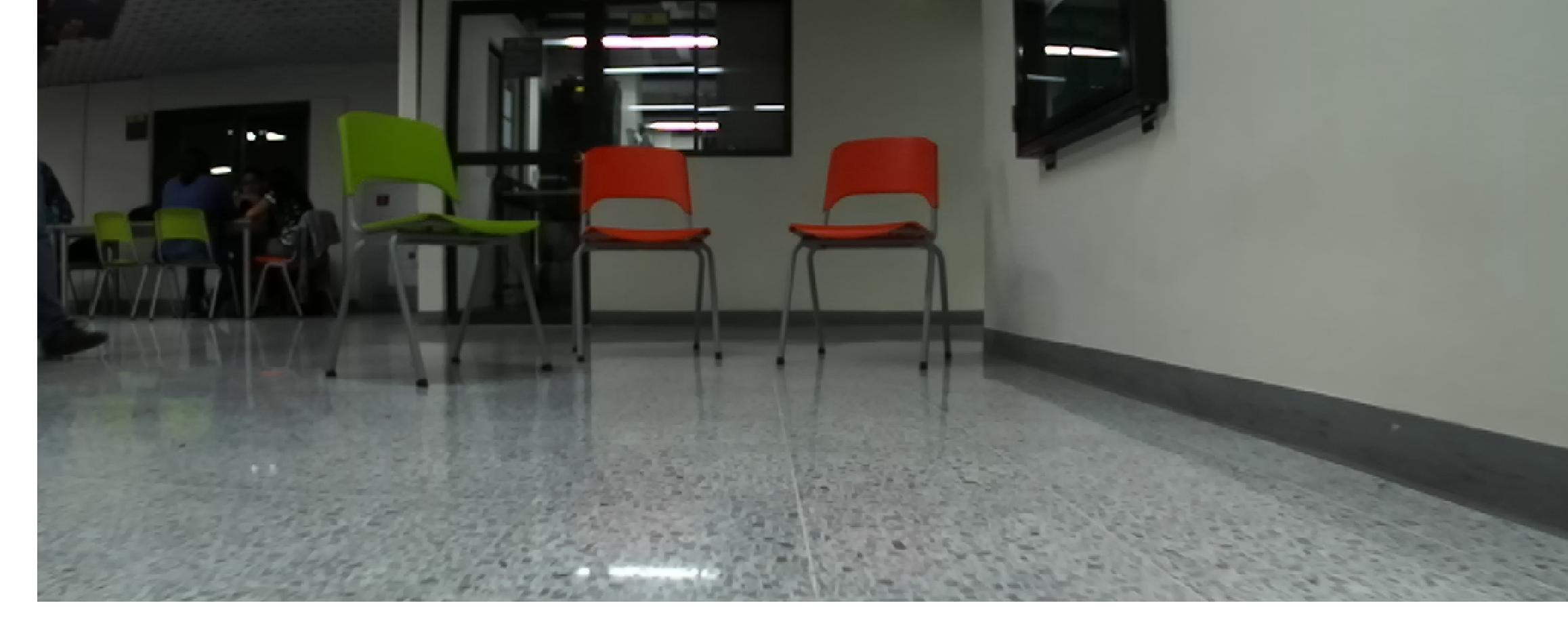}
 	   
 	   \includegraphics[trim={0cm 0 0.6cm 0},clip,scale=.25]{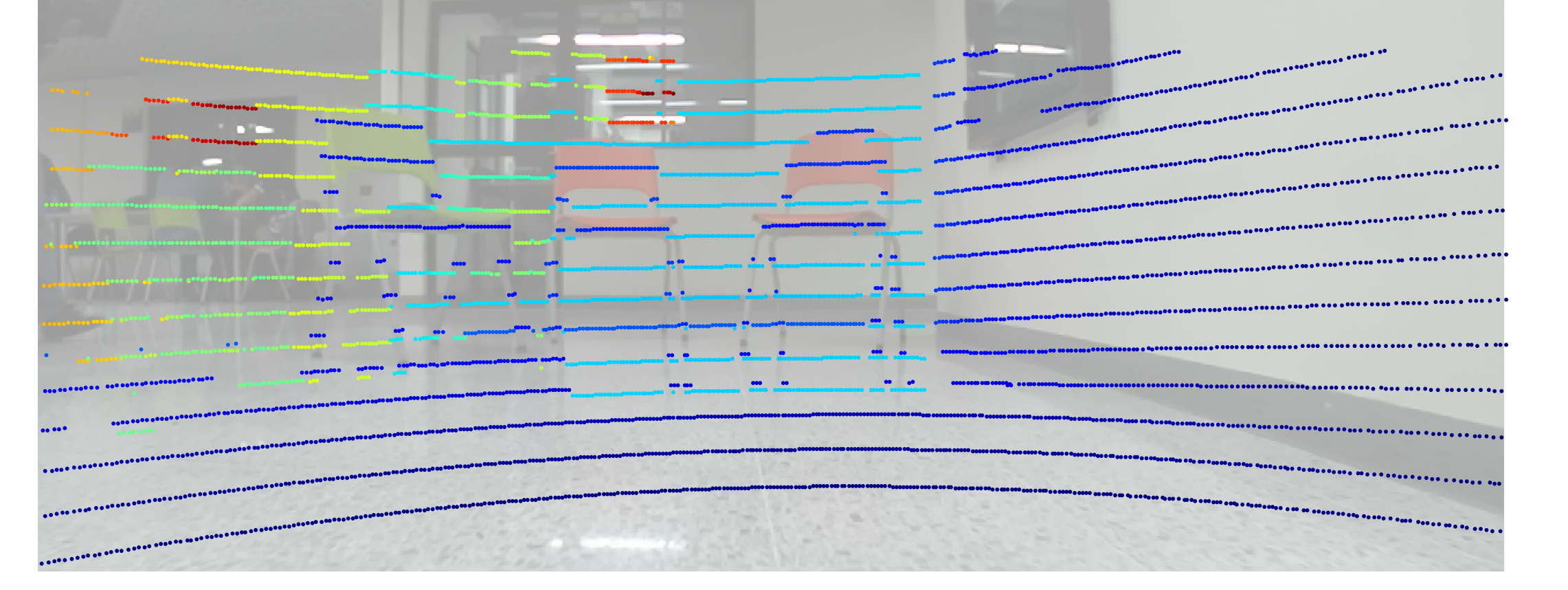}

		\includegraphics[trim={0.7cm 0 0cm 0},clip,scale=.179]{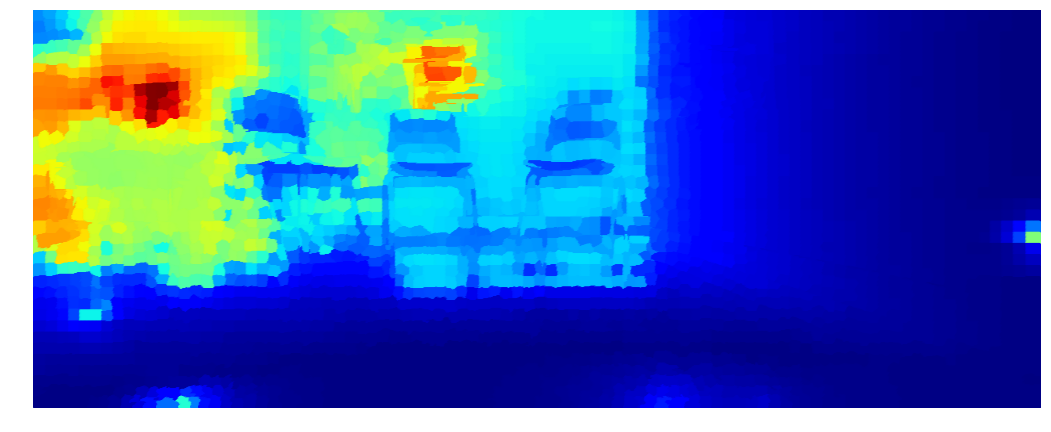}
		
	\caption{Indoors: LiDAR super-resolution.
Creating dense point clouds from sparse raw measurements and colour.
From top to bottom: RGB image, raw depth map and predicted depth.}
	\label{FIG:9}
\end{figure}

\begin{figure}
	\centering

		\includegraphics[trim={0.8cm 0 0.3cm 0},clip,scale=.238]{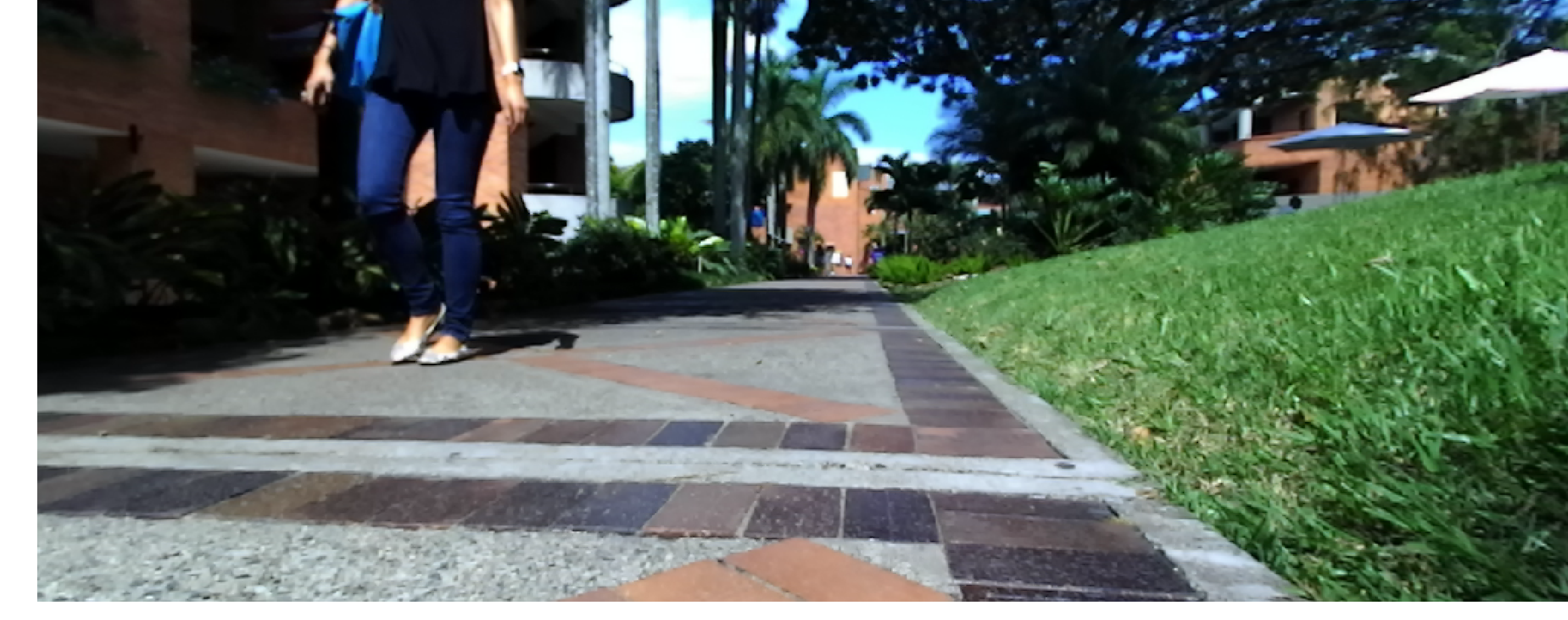}

		\includegraphics[trim={0cm 0 0.6cm 0},clip,scale=.25]{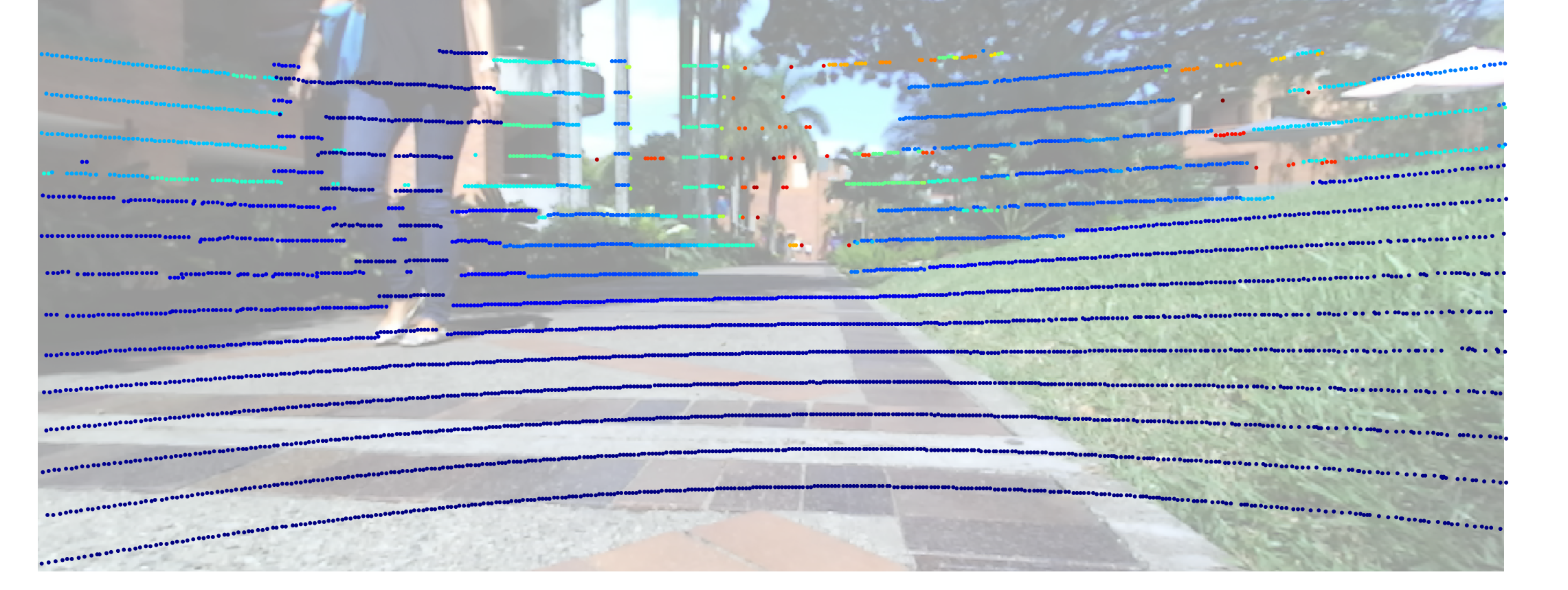}

		\includegraphics[trim={0.7cm 0 0cm 0},clip,scale=.1826]{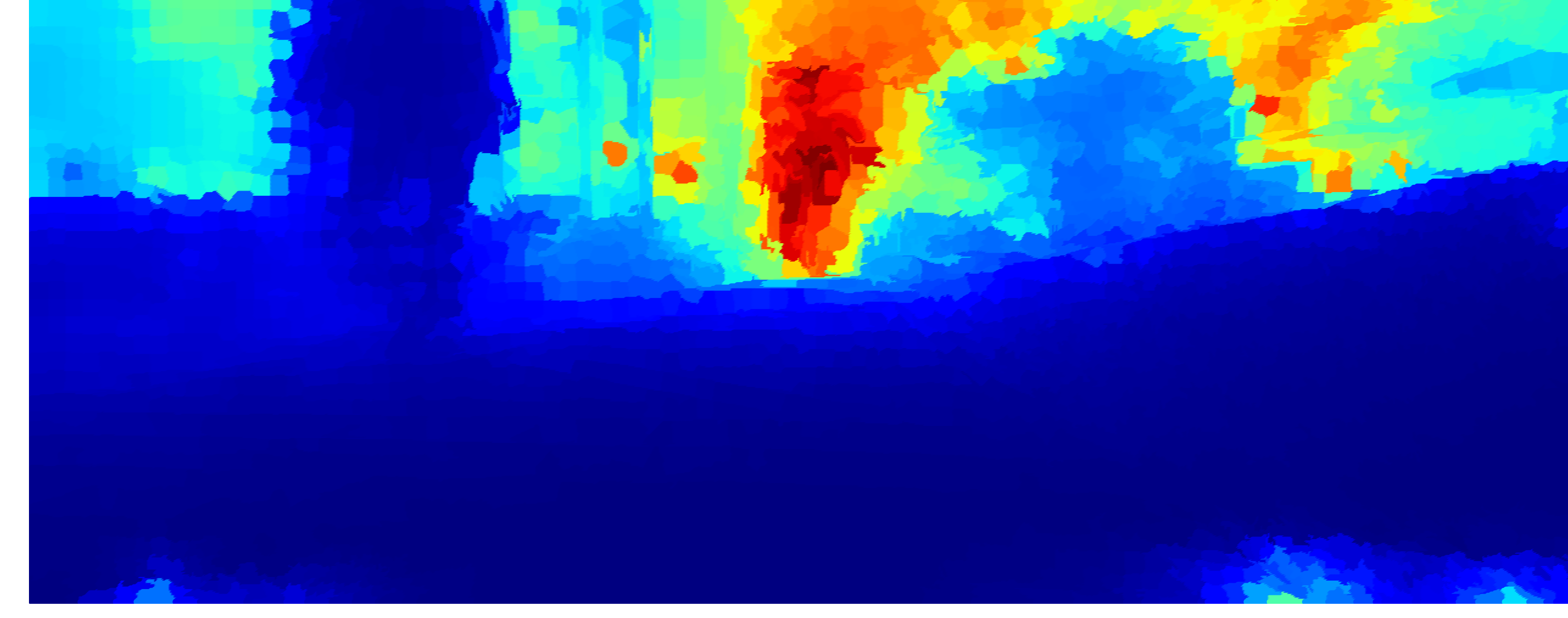}	
		
	\caption{Outdoors: LiDAR super-resolution.
Creating dense point clouds from sparse raw measurements and colour.
From top to bottom: RGB image, raw depth map and predicted depth.
}
	\label{FIG:10}
\end{figure}

\begin{figure}
	\centering

		\includegraphics[trim={0.8cm 0 0.3cm 0},clip,scale=.238]{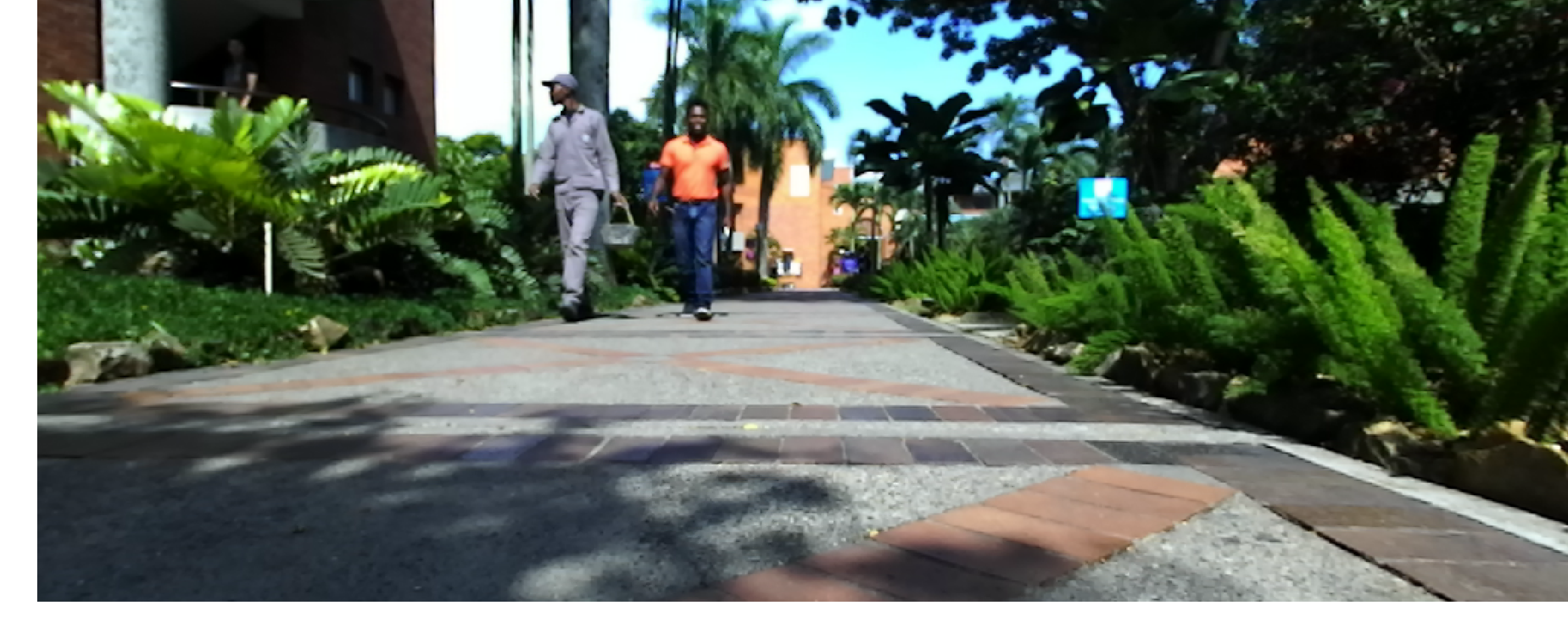}
		
		\includegraphics[trim={0cm 0 0.6cm 0},clip,scale=.25]{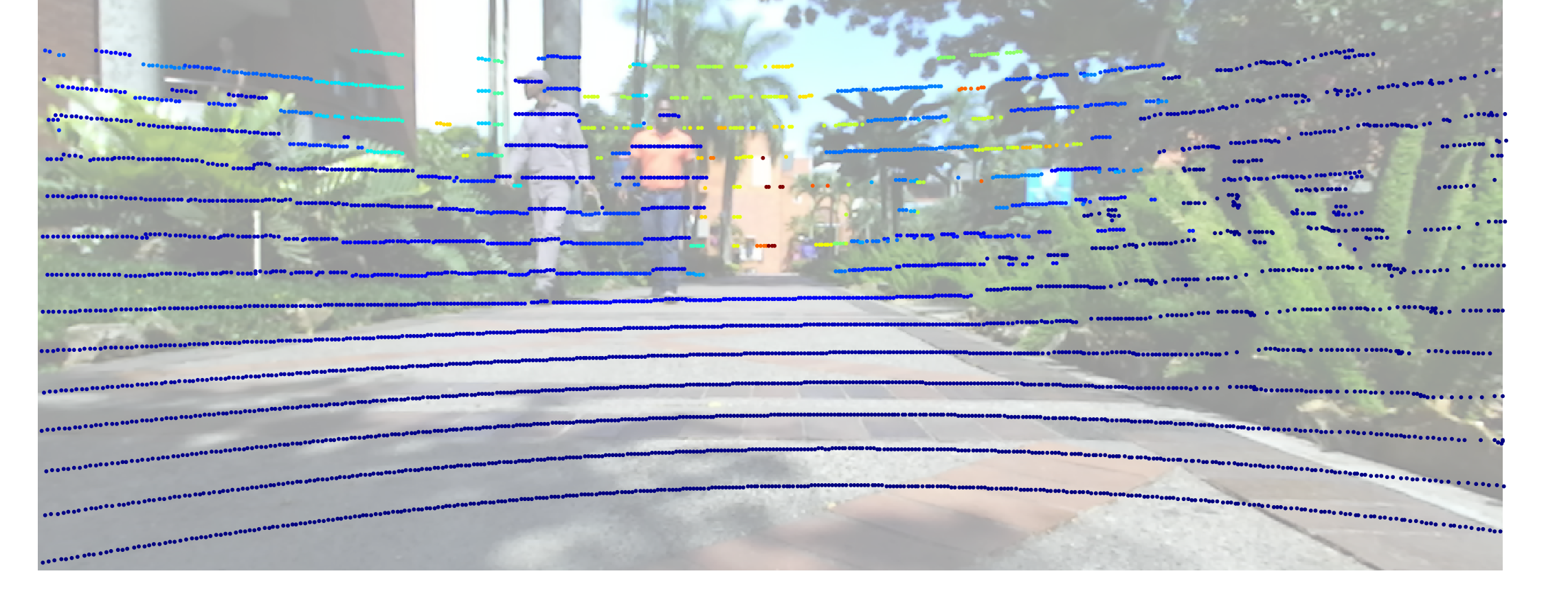}
		
		\includegraphics[trim={0.7cm 0 0cm 0},clip,scale=.1826]{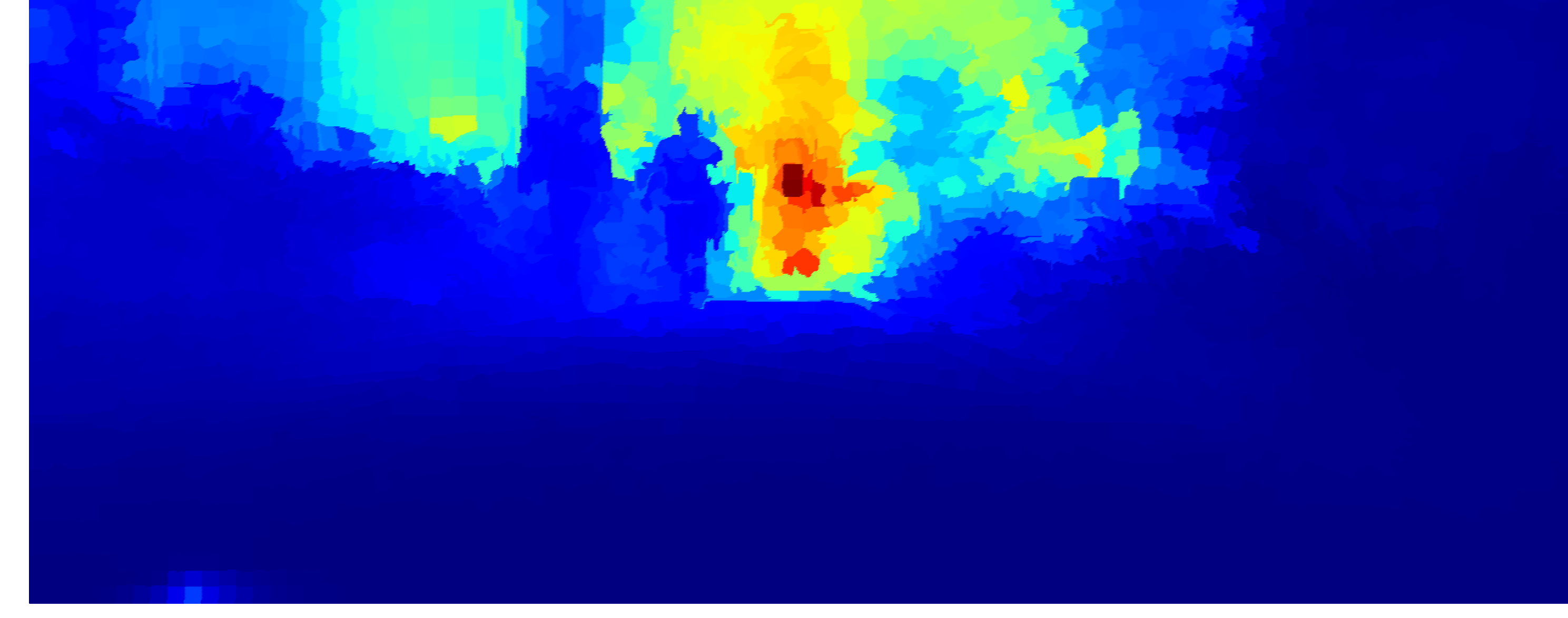}
		
	\caption{Outdoors: LiDAR super-resolution.
Creating dense point clouds from sparse raw measurements and colour.
From top to bottom: RGB image, raw depth map and predicted depth.
}
	\label{FIG:11}
\end{figure}

Notably, the proposed algorithm is able to estimate a dense depth map of indoor and outdoor environments using colour and sparse depth data.
The experimental results are shown in \textcolor{blue}{Fig.~\ref{FIG:9}}, \textcolor{blue}{Fig.~\ref{FIG:10}} and \textcolor{blue}{Fig.~\ref{FIG:11}}.
 Dark red indicates farther distances and dark blue indicates closer distances.

Despite  the  lower  number  of LiDAR  channels, the  proposed method has provided accurate depth information even under challenging outdoor conditions,  as  shown in \textcolor{blue}{Fig.~\ref{FIG:11}}.
In this scene there is lot of variability in terms of the light and shadows generated by the environment and the weather itself.

After a close look at \textcolor{blue}{Fig.~\ref{FIG:9}}, \textcolor{blue}{Fig.~\ref{FIG:10}} and \textcolor{blue}{Fig.~\ref{FIG:11}}, it is noticeable that no depth observations from the LiDAR are available at the top and bottom locations of the colour image. After inference, the depth estimates, shown in the bottom images, at the above locations are consistent with the information provided by the image. We can conclude that the framework proposed here works reliably for the depth prediction task. Additionally, it also solves the depth completion problem, as it is able to deal with highly sparse input point clouds projected onto the image space.

\section{Conclusions and future work}
\label{CFW}

In this paper, we described an innovative approach to fuse information from different sensor modalities, e.g., cameras and LiDAR, in order to probabilistically estimate a dense point cloud.

Our approach achieves good performance in single image depth map prediction on the popular KITTI dataset. It is able to predict detailed depth maps on thin and distant objects.
It also reasonably estimates the depth in parts of the image in which there is no ground-truth available for supervised learning.

The qualitative and quantitative results presented here suggest that our system provides 2D depth maps of reasonable quality, which depends on the density of the laser measurements and the number of superpixels selected. We believe that this method opens up an important avenue for research into multi-sensor fusion and the more general 3D perception problems, which might benefit substantially from sparse depth samples. 

There are various variants of CRFs, each with its own pros and cons. A Higher-order CRF, for example, considers interactions between more than two variables at a time. This can lead to more accurate predictions. A Dynamic CRF is another type of CRF that is designed to work with sequences of data, such as time series or videos. This can be useful for depth estimation using Camera-LiDAR fusion, as it can take into account temporal information in the data. However, both Higher order and Dynamic CRFs require more computational resources and are more difficult to train than pair-wise, also known as Markov CRFs, which only consider interactions between adjacent variables. Since our work focuses on robotics applications, in which computational performance is a hard constraint, instead of trying to capture long-term dependencies in the data, we leverage the potential of modeling both appearance and geometric constraints from multimodal sensor observations. A promising future research direction may tackle the challenges of developing efficient and real-time inference solutions for higher order CRF’s in the context of multi-modal depth estimation solutions. Another promising research venue is the parallel implementation of our method, which will bring benefits when deploying it in real robotic platforms. Additionally, the inclusion of other sensor modalities, such as radar \cite{9340998, long2021radar}, could be explored as a way of improving our system's robustness to challenging environmental conditions.

\section{Acknowledgment}
This work was supported by Universidad Autónoma de Occidente (UAO).
The authors would like to thank the Research incubator in robotics and autonomous systems (RAS), the Research group on remote and distributed control systems (GITCoD) at UAO, Walter Mayor, Nicolas Llanos Neuta and Juan Carlos Perafán for their feedback and helpful discussions.

\printcredits

\bibliographystyle{cas-model2-names}

\bibliography{references}

\newpage

\bio{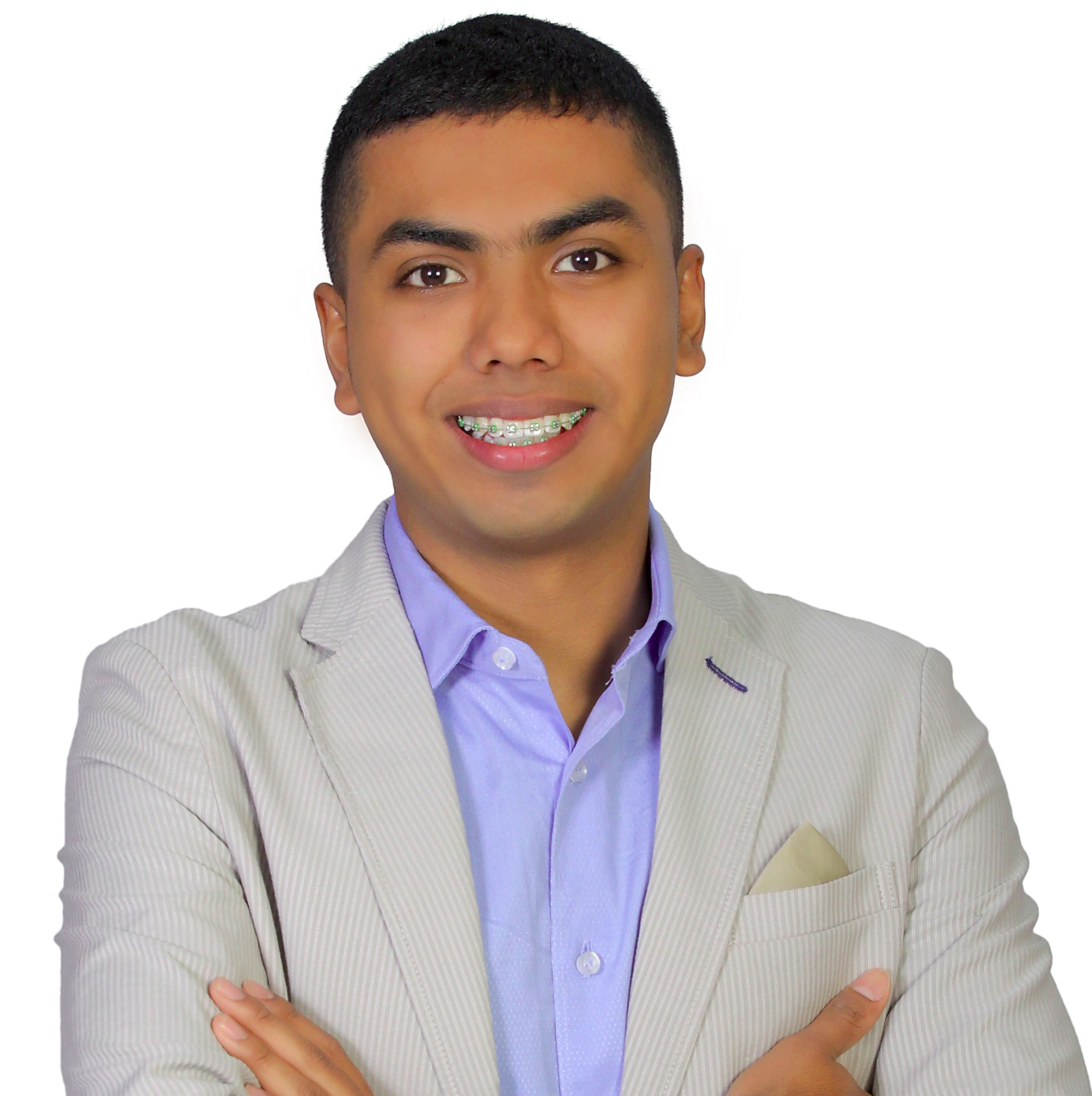}
Johan S. Obando-Ceron received the B.S. degree in mechatronics engineering from the Universidad Autónoma de Occidente (Autonomous University of the West), Colombia,   in   2017, where he worked for three years as a research assistant. His current research interests include reinforcement learning, as  well  as  statistics and optimization theory and its applications.

\endbio

\bio{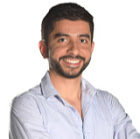}
Victor Romero-Cano received a B.S. degree in mechatronic engineering from Universidad Autónoma de Occidente, Colombia, in 2007, an MSc degree in electrical engineering from the Universidade de Sao Paulo, Brazil, in 2010, and a PhD from the Australian Centre for Field Robotics, University of Sydney, Australia, in 2015. He joined INRIA as a postdoctoral research associate within the CHROMA team in 2016. Since 2017, he has been a lecturer and researcher at the Universidad Autónoma de Occidente. His research interests include multi-sensor robotic perception and machine learning.
\endbio

\bio{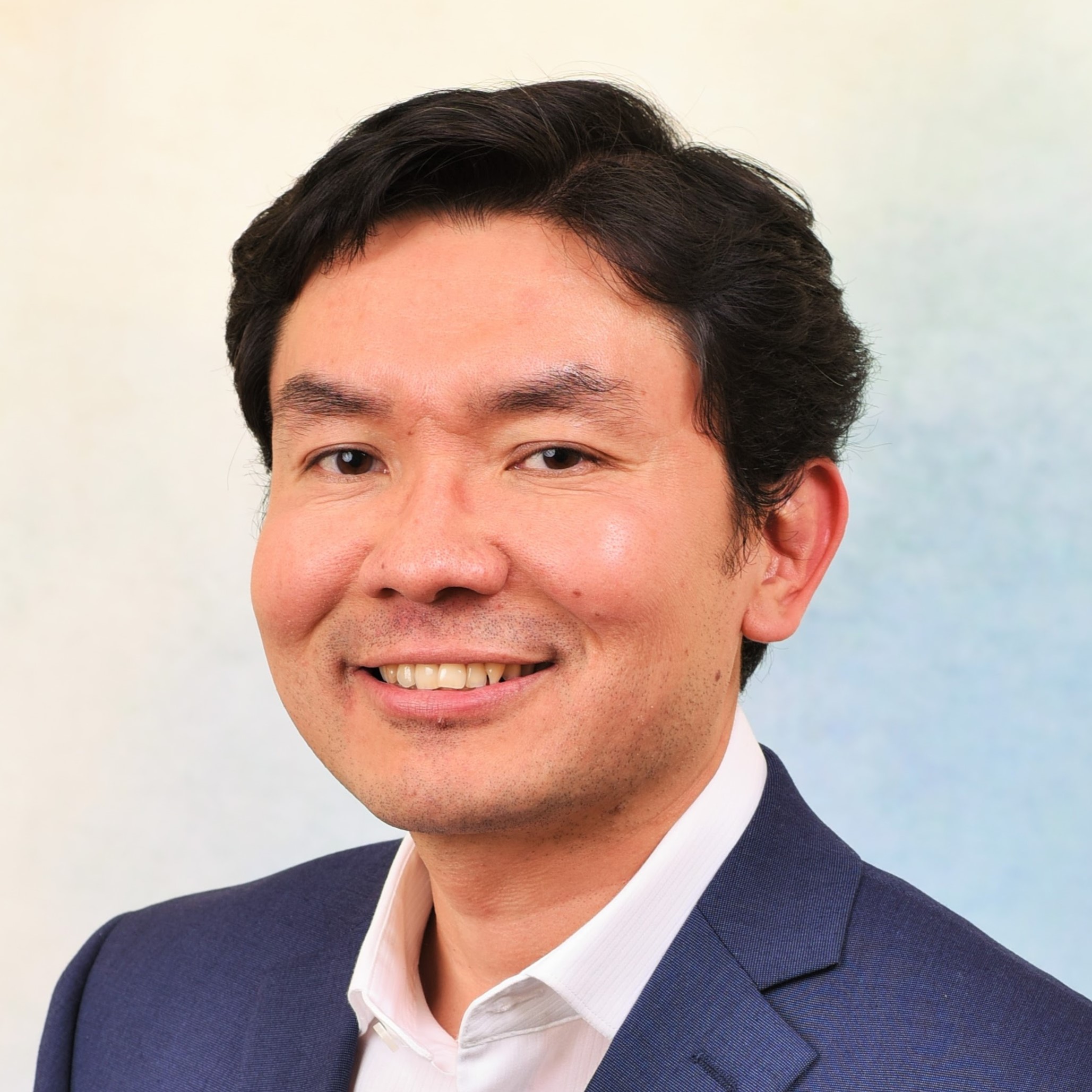}
Sildomar Monteiro received an MSc degree from the Instituto Tecnológico de Aeronáutica, Brazil, and a PhD from the Tokyo Institute of Technology, Japan, in 2007. He is a research scientist working on AI, machine learning, computer vision, and robotics for autonomous systems. He enjoys leading and mentoring teams to develop rigorous approaches for challenging real-world problems.
\endbio

\end{document}